%% file: ms.tex
\documentclass[acmsmall]{acmart}

\AtBeginDocument{%
  \providecommand\BibTeX{{%
    \normalfont B\kern-0.5em{\scshape i\kern-0.25em b}\kern-0.8em\TeX}}}
    
\usepackage{graphicx}

\usepackage{amsmath}
\usepackage{tabularx}
\usepackage{booktabs}
\usepackage{scrtime}
\usepackage[dont-mess-around]{fnpct}
\usepackage{etoolbox}

\setcopyright{acmcopyright}
\copyrightyear{2021}
\acmYear{2021}
\acmDOI{10.1145/1122445.1122456}

\acmJournal{THRI}
\acmVolume{1}
\acmNumber{1}
\acmArticle{1}
\acmMonth{1}

\begin{document}

\title{In the Arms of a Robot: Designing Autonomous Hugging Robots with Intra-Hug Gestures}
\renewcommand{\shorttitle}{Designing Autonomous Hugging Robots with Intra-Hug Gestures}

\author{Alexis E. Block}
\email{alexis@is.mpg.de}
\orcid{0000-0001-9841-0769}
\affiliation{%
  \institution{Max Planck Institute for Intelligent Systems and ETH Z\"{u}rich}
  \city{Stuttgart}
  \country{Germany}
}

\author{Hasti Seifi}
\orcid{/0000-0001-6437-0463}
\affiliation{%
  \institution{University of Copenhagen}
  \city{Copenhagen}
  \country{Denmark}}
\email{hs@di.ku.dk}

\author{Otmar Hilliges}
\orcid{0000-0002-5068-3474}
\affiliation{%
  \institution{ETH Z\"{u}rich}
  \city{Z\"{u}rich}
  \country{Switzerland}}
  \email{otmar.hilliges@inf.ethz.ch}

\author{Roger Gassert}
\orcid{0000-0002-6373-8518}
\affiliation{%
 \institution{ETH Z\"{u}rich}
 \city{Z\"{u}rich}
 \country{Switzerland}}
 \email{roger.gassert@hest.ethz.ch}

\author{Katherine J. Kuchenbecker}
\orcid{0000-0002-5004-0313}
\affiliation{%
  \institution{Max Planck Institute for Intelligent Systems}
  \city{Stuttgart}
  \country{Germany}}
  \email{kjk@is.mpg.de}

\input{./sections/00_abstract}





\maketitle

\input{./sections/01_introduction}
\input{./sections/02_related_work}
\input{./sections/03_design_guidelines}
\input{./sections/04_materials_and_methods}

\input{./sections/05_results}
\input{./sections/table_huggiebot}
\input{./sections/06_changes_to_platform}
\input{./sections/07_validation}

\input{./sections/08_results2}
\input{./sections/09_discussion}
\input{./sections/10_conclusion}

\begin{acks}
This work is partially supported by the Max Planck ETH Center for Learning Systems and the IEEE RAS Technical Committee on Haptics. The authors thank Sammy Christen, Felix Gr\"{u}ninger, Joey Burns, Bernard Javot, Ilona Jacobi, Meike Pech, Natalia Egana Rosa, Ravali Gourishetti, Neha Thomas, Natalia Sanchez-Tamayo, Mayumi Mohan, Christoph Ricklin, and Kinova Robotics for supporting various aspects of this research project. 
\end{acks}

\bibliographystyle{ACM-Reference-Format}
\bibliography{sample-base}

\input{./sections/11_appendix}


\end{document}

%% file: sections/00_abstract.tex
\begin{abstract}
Hugs are complex affective interactions that often include gestures like squeezes. We present six new guidelines for designing interactive hugging robots, which we validate through two studies with our custom robot. To achieve autonomy, we investigated robot responses to four human intra-hug gestures: holding, rubbing, patting, and squeezing. Thirty-two users each exchanged and rated sixteen hugs with an experimenter-controlled HuggieBot 2.0. The robot's inflated torso's microphone and pressure sensor collected data of the subjects' demonstrations that were used to develop a perceptual algorithm that classifies user actions with 88\% accuracy. Users enjoyed robot squeezes, regardless of their performed action, they valued variety in the robot response, and they appreciated robot-initiated intra-hug gestures. From average user ratings, we created a probabilistic behavior algorithm that chooses robot responses in real time. We implemented improvements to the robot platform to create HuggieBot 3.0 and then validated its gesture perception system and behavior algorithm with sixteen users. The robot's responses and proactive gestures were greatly enjoyed. Users found the robot more natural, enjoyable, and intelligent in the last phase of the experiment than in the first. After the study, they felt more understood by the robot and thought robots were nicer to hug. 
\end{abstract}

%% file: sections/01_introduction.tex
\section{Introduction}
\label{Introduction}

\begin{figure}
  \includegraphics[trim = {0cm 0cm 0cm 0cm}, clip, width=\textwidth]{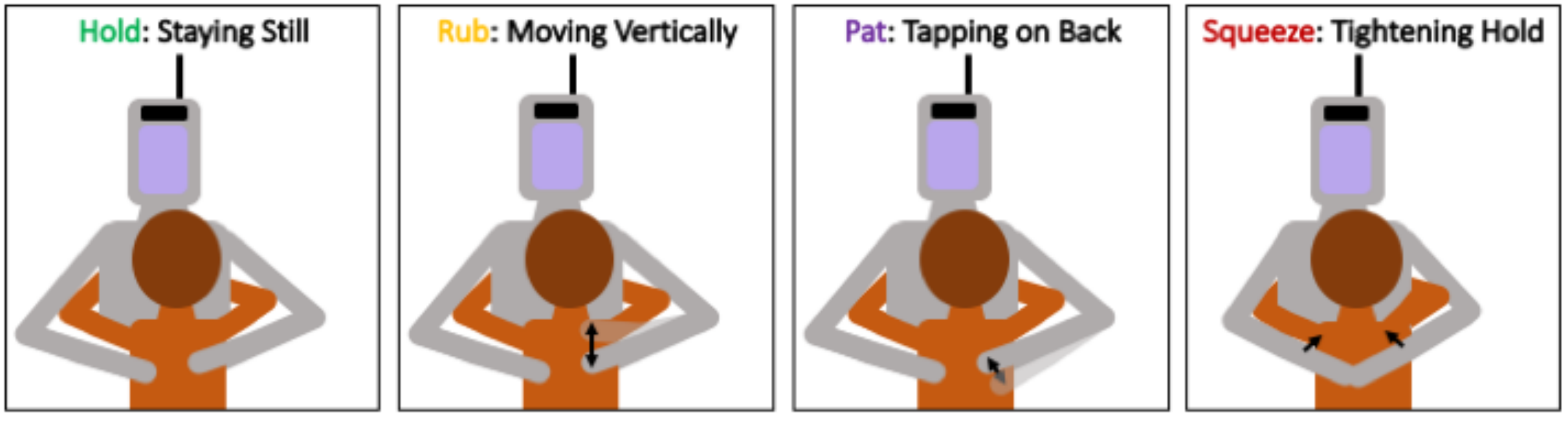}
  \caption{The four intra-hug gestures that our hugging robot, HuggieBot, can perform, either in response to a user action or proactively when it does not detect any user actions. Despite their importance during prolonged hugs between humans, no prior hugging robot has been able to detect and respond to intra-hug gestures.}
  \Description{}
  \label{fig:teaser}
\end{figure}
From the moment we are born, social touch affects our future ability to function well in society. Infants who are held by their mothers for two hours after they are born have better interactions with their mothers and are better at handling stress \cite{Uvnas-Moberg2014}. In such a close, positive relationship, the hormone oxytocin is released when the two partners see, hear, or even think of each other. In turn, this release bonds them even more closely and improves positive human relationships. A similar positive calming response can be evoked during an embrace, a massage, or other stroking of the skin, often called deep pressure touch \cite{SqueezeMachine}. Hugs that last more than three seconds often include intra-hug gestures, like squeezes and rubs (Fig.~\ref{fig:teaser}), that create a close physical exchange between the two participants and confer additional benefits from the increased deep pressure touch \cite{3secondhug}. 

Not everyone is fortunate enough to have close, positive relationships with people around them. If a similar effect can be achieved through a robotic embrace, this helpful touch can benefit people who otherwise would not be able to experience hugs. The broader goal of this research project is to \textit{provide an embodied affective robot that can supplement human hugs in situations when requesting this form of comfort from others is difficult or impossible}. Many common examples where people lack access to human hugs stem from long-term physical separation. During the current global pandemic of COVID-19, some family members have been unable to come into close contact with each other for more than one year, and the effects are showing. Lack of social touch can be detrimental to both our physical and mental health \cite{DepressedTeensFromSocialMedia, InternetAndDepression, SocialTouchDevelopment}. 

Accurately replicating a human hug is a difficult problem because it requires \textit{real-time adaptation to a wide variety of users}, \textit{close physical contact}, and \textit{quick, natural responses to intra-hug gestures performed by the user}. In the past, researchers have avoided tackling these challenges by providing a huggable device that does not actively hug the user back, thereby entirely avoiding the challenges of reciprocating a hug \cite{HuggablePillowWithPhone, Hugvie, TheHuggable, disalvo2003hug}. Others have chosen to create robots that hug in a ``one-size-fits-most'' model \cite{HugShirt, hedayati2019hugbot, shiomi2017hug, shiomi2017robot, tsetserukou2010haptihug}. Another set of researchers adjusted the robot to each specific user prior to experimentation, thereby avoiding the challenge of real-time adaptation \cite{block2019softness, block2018emotionally}. \citet{TheSixHugCommandments} recently introduced HuggieBot 2.0 as the first robot that uses visual and haptic perception to deliver closed-loop hugging that adapts to the circumference of the user and their preferred hug timing; however, this robot could not perceive or respond to intra-hug gestures, and user testing revealed other limitations. Section~\ref{RelatedWork} further details prior research in this domain, discussing the importance of social touch between humans, summarizing different forms of technology-mediated social touch, reviewing the state of the art for social touch in human-robot interaction (HRI), and summarizing scientific approaches to evaluating user experience. 

We accept and build upon the six design guidelines for hugging robots previously presented by Block et al. \cite{TheSixHugCommandments}: a good hugging robot should: (G1) be soft, (G2) be warm, (G3) be sized similar to an adult human, (G4) visually perceive and react to an approaching user, (G5) autonomously adapt its embrace to the size and position of the user's body, and (G6) reliably detect and react to a user's desire to be released from a hug regardless of their arm positions. Section~\ref{DesignGuidelines} presents a \textbf{refined version of G4} plus \textbf{five additional design guidelines} for autonomous hugging robots that we derived from the findings of the two studies reported in this paper. Our new presented guidelines state that a good interactive hugging robot should: (refined G4) autonomously initiate a hug in a consensual and synchronous manner, (G7) adapt to the user's height, (G8) perceive intra-hug gestures in real time, (G9) respond quickly to detected gestures, (G10) slightly vary its response to each detected intra-hug gesture, and (G11) occasionally perform proactive affective gestures during hugs.  Section~\ref{DesignGuidelines} also defines the goal of this research and outlines our reasoning for the process we followed.

To create a robot that can autonomously deliver pleasant, natural-feeling hugs, we first conducted a Wizard-of-Oz user study (action-response elicitation study) with HuggieBot 2.0 to collect data on intra-hug gestures; Section~\ref{UserStudyMethods} explains the methods, and Section~\ref{Results} presents and briefly discusses the results. As described in Section~\ref{Improvements}, we improved several aspects of the platform's hardware and software based on user feedback from this study as well as a prior evaluation of HuggieBot 2.0. The collected data were then used to develop a perception system and a behavioral response algorithm for the updated version of our platform, HuggieBot 3.0. Specifically, Section~\ref{Detection} explains how we analyzed the microphone and pressure sensor data collected from our novel inflated robot torso (HuggieChest) as 32 diverse users performed four distinct intra-hug gestures. Our developed machine-learning methods quickly detect and reliably classify these different gestures. Based on the 32 users' ratings of the different robot responses, we developed a probabilistic behavior algorithm to determine which action the robot should perform in response to a user gesture; it is described in Section~\ref{Response}. Rather than maximizing user acceptance for each robot gesture, which would result in the robot only squeezing the user, our behavior algorithm balances exploration and exploitation \cite{explorationexploitation} to create a natural, spontaneous robot that provides comforting hugs.

As detailed in Section~\ref{Validation}, we then ran a follow-up study with sixteen new users to test our detection and classification system's real-world accuracy and evaluate the user acceptance of our robot behavior algorithm (validation study). Section~\ref{Results2} shares the results of this study, which show that HuggieBot 3.0 is the \textit{first fully autonomous human-sized hugging robot that recognizes and responds to the user's intra-hug gestures.} Section~\ref{Discussion} discusses the results of the validation study in the context of our six new hugging design guidelines, and it also addresses the limitations of our approach, such as the lab setting and the number of participants in the validation study. Finally, Section~\ref{Conclusion} provides a summary of this article and shows various avenues for future work on this topic.  We believe the presented design guidelines can be expanded beyond HuggieBot to give a wide range of companion robots the ability to exchange high-quality hugs with their human users.


%% file: sections/02_related_work.tex
\section{Related Work}
\label{RelatedWork}
\subsection{Social Touch Between Humans}
\label{subsec:socialtouchpeople}
People interact with each other through various kinds of social touch that change across the lifetime in order to foster a sense of community, strengthen relationships, and build trust \cite{SocialTouchDevelopment}. The first form of affective, social touch usually comes from a mother or other relative to help soothe an infant \cite{Uvnas-Moberg2014}. In addition to calming the child down, this helpful touch (cradling, squeezing, hugging, kissing, stroking, etc.) strengthens the bond between adult and child and increases the body's production of oxytocin in both. An increase in oxytocin provides a host of benefits, including a greater tolerance for pain and stress \cite{LowerCortisol}.

Over time as they grow, rather than relying on a parent for comfort, humans learn to self-soothe in similar ways, such as holding themselves, rubbing their arms, and wrapping themselves tightly in a warm blanket. These methods work because they are reminiscent of deep pressure touch, which is the kind of touch one receives when hugging someone or touching them firmly. It has a calming effect and has been shown to alleviate stress and anxiety dramatically and lower heart rates and blood pressure \cite{Edelson1999}. 

As humans age further, researchers have found that the areas a person is allowed to touch on another's body are directly correlated with the strength of the relationship between the two people \cite{suvilehto2015topography}. The closer two people are emotionally to each other, the more areas they are allowed to contact for social touch. Appropriate location of touch was of particular importance to our research, as Block et al. found users to be highly sensitive to the hand placement of their hugging robot; placing the robot's arms either too low or too high on the body was not appreciated \cite{TheSixHugCommandments}.

In addition to the location of social touch, another critical element to consider during social touch between humans is the applied contact strength. Too little pressure can sometimes create a disingenuous impression or not provide enough emotional support, while too much pressure can hurt your partner. Researchers studied the physiological responses (heart rate and R-R interval) of infants being held with different levels of tightness and by people with varying relationships to the child \cite{Yoshida2020}. Infants responded best when being held with a medium amount of contact pressure by a parent. This study identifies a Goldilocks zone where the pressure is not too low and not too high, but ``just right.'' This pressure zone may vary across people and can even change for a single person depending on how they feel and how much support they currently need (more support may require more applied pressure).

A common way to appropriately provide a person with deep pressure touch is through a hug, where both participants wrap their arms around the other person's torso. During prolonged hugs (lasting more than three seconds), the two hugging partners rarely remain stationary in the embrace \cite{3secondhug}. Common intra-hug gestures like stroking/rubbing, patting, and squeezing provide comfort and help the receiver feel emotionally supported  \cite{Waal2019}. Although few other investigations have explored intra-hug gestures, this research inspired us to create a robot that can detect and respond to such gestures during a hug to more closely mimic humans, so that it can eventually provide better emotional support to its users.

\subsection{Technology-Mediated Social Touch}
Because physical contact is not always possible, some researchers look for ways to strengthen relationships between emotionally close people who are separated by a distance. As will be explained in the following paragraphs, this gap can be bridged by technology that aims to transmit social touch, e.g., \cite{Pakanen2014, disalvo2003hug}. \citet{Huisman2017} provides a comprehensive review of social-touch technologies. The developed devices typically work in pairs, where users can send signals to each other to let their partner know they are thinking of them, e.g., \cite{HeyBracelet, CoupleVibe, FriendshipLamp}. The signal the other user receives indicates their partner is sending them some contact, but it may not accurately replicate the sender's desired intent (e.g., vibration output may be used to represent a squeezing input). Such objects typically fall into two categories that we discuss in more detail below: wearables and comfort objects.

\subsubsection{Wearables}
Carrying your loved one with you wherever you go can be an ideal solution for those in a long-distance relationship. In this subsection, we discuss previous work on developing wearable technology to help physically separated loved ones feel emotionally close to each other, e.g., \cite{Pakanen2014, HeartBeatAppleWatch, duvall2016active}. A common version of this wearable technology occurs in the form of a bracelet. The Squeezy bracelet \cite{Pakanen2014} can be paired with a mobile phone and allows the user to receive haptic messages. The Hey Bracelet \cite{HeyBracelet} works in pairs and allows friends to send each other haptic signals by connecting them with their phones to Bluetooth. When one user taps on their Hey Bracelet, the other one will squeeze and vibrate to let the wearer know their partner is sending them a hug. Similar capabilities are also possible with Apple Watches, where you can send a heartbeat to anyone via iMessage \cite{HeartBeatAppleWatch}. Another bracelet developed is CoupleVIBE \cite{CoupleVibe}, which sends a vibrotactile cue to a partner to signal that their long-distance significant other has arrived to or departed from a frequented location. 

The HaptiHug is another kind of wearable for social touch \cite{tsetserukou2010haptihug}; it is similar to a belt and features soft hands made from rubber-sponge material. The creators developed an animated version of a hug and integrated it into the online game Second Life. Real-world users can connect and interact in this virtual world. By wearing the HaptiHug while playing the game, users can feel squeezed and have pressure applied on their back when their virtual characters hug each other, thus making the experience more immersive. HaptiHug provides three pre-programmed levels of hug intensity from which users can choose.

Inflatable and weighted vests have been used to provide children on the autism spectrum with deep pressure touch \cite{duvall2016active}. While they do succeed in delivering this beneficial touch, the inflatable vest's loud pumps are conspicuous and distracting. In contrast, the weighted vests must be removed and replaced frequently for the wearer to feel the benefit. An additional potential drawback of the inflatable vest is that someone can remotely operate it at any time without warning to the child, who could have no idea when, why, or from where the hug is coming. 

Another kind of wearable technology for mediated social touch is the Hug Shirt \cite{HugShirt}, which has embedded sensors and actuators. The sender and the receiver each wear a Hug Shirt, and the system aims to create the sensation of physical contact between the wearers. The sensors capture the sender's contact location, strength, warmth, and heartbeat, and the actuators attempt to recreate these sensations on the receiver's shirt through heating, vibration, and inflation. The shirts send messages to each other via Bluetooth by connecting to the mobile phones of the users.

A final wearable we will discuss is the Huggy Pajama \cite{teh2008huggypajama}. These pajamas are meant to be a hugging communication system between children and parents. A parent hugs a doll embedded with sensors, and the child wearing the pajamas feels virtually hugged. The pajamas are actuated by air inflation with a compressor located outside of the pajamas. Unfortunately, this compressor is loud and can be disruptive to a child trying to sleep, though the provided compression was reported to be enjoyable. In this example and all others discussed above, the wearable was not capable of detecting, classifying, or responding to intra-hug gestures. 

\subsubsection{Comfort Objects} 
The idea of technology-mediated social touch is so compelling that people are purchasing ``Friendship Lamps'' \cite{FriendshipLamp} even without research supporting their claimed benefits. These lamps work in pairs. When one user turns on theirs, the partner's lamp lights up. Touching either lamp causes both lamps to change colors to indicate to the partner that the other person is thinking of them. Another group of researchers took the idea of a tele-lamp one step further by giving it an anthropomorphic shape intended for affective interaction \cite{Angelini2015}. This lamp has a face, can change colors, and displays different emotions to reflect a distant loved one's emotional state. A user can change the lamp's emotional state by performing gestures on it, for example kissing the lamp. These researchers also share that while couples can use this lamp for long-distance relationships, it can also be used alone as a single companion. 

Another group of researchers developed Hugvie \cite{Hugvie, Nakanishi2020}, a small pillow with the approximate shape of a person. The pillow's head contains a small pocket in which a user can place a cell phone. Users then hug the pillow while talking on the phone to their far-away loved one. Researchers found that using this technology can help reduce the hugger's anxiety \cite{Hugvie, Nakanishi2020}. They also found the interpersonal touch from using Hugvie can help improve the hugger's impression of a third person based on hearsay information given by the remote partner. However, this system cannot detect, classify, or respond to any intra-hug gestures. 

Similar to the friendship lamps, other researchers developed a plush pillow, the Macaron \cite{Nunez2017}, that uses infrared photo-reflective sensors to detect when a user is hugging it. It then sends a message over Bluetooth to the partner pillow, which lights up with blue-colored LEDs and blinks to indicate the intensity with which the partner hugged the other Macaron.  Likewise, another group of researchers created The Hug \cite{disalvo2003hug}, a pillow whose shape is derived from a child wrapping its legs around an adult. The Hug also works in pairs. When one user hugs or strokes their Hug, the partner Hug will light up, vibrate, jingle, and heat up to indicate that someone is sending a hug. 

\subsection{Human-Robot Hugging}
\label{subsec:human-robot-social-touch}
Social touch occurs in many contexts. Several researchers have devoted effort to enabling robots to shake hands with humans as naturally as humans do with each other, e.g., \cite{Arns2017, Wang2009}. Other researchers have worked to allow humans to connect with robots in a more light-hearted and playful manner through high-fives and hand-clapping games \cite{Fitter2016,Fitter19-IJSR-Clap,Fitter20-JNER-Exercising}. All these interactions feature ways to help humans and robots interact with each other both socially and physically. This interest in enabling social-physical human-robot touch is not unique; many researchers have taken different approaches to this goal. We will focus on the most common approaches that are relevant to our research on HuggieBot involving social touch through human-robot hugs. 


Perhaps the most well-known of human-robot hugs is Temple Grandin's Squeeze Machine \cite{SqueezeMachine}. Though not technically a robot, this machine applies lateral deep touch pressure by squeezing the user between two foam panels. This machine is user-operated, so it does not require an additional human partner. The user directly sets the duration and pressure of the squeeze they receive.

The Huggable is a small robotic teddy bear companion meant to accompany small children during extended stays in the hospital \cite{TheHuggable}. The robot can detect where and how the child is touching it, and it can move its head and arms in response. The Huggable has cameras, microphones, and speakers, and it can record the child using it and send helpful information to a remote caregiver. Due to its small form factor, this robot is huggable, but it cannot actively hug the user back.

Shiomi et al. created a life-size teddy bear robot with back-driveable motors at the elbows that was evaluated in two Wizard-of-Oz experiments \cite{shiomi2017hug, shiomi2017robot}. While the system currently requires a human operator, the hug itself is meant to come from the robot, not another user; thus, we consider it a human-robot hug rather than a technology-mediated hug. This floor-sitting robot introduces itself to users before asking for a hug. The user must then crawl toward the robot to hug it. The creators found that hugging this robot caused users to engage in more self-disclosure and be more willing to donate money to charitable causes. 

HugBot is a large panda bear stuffed animal \cite{hedayati2019hugbot} similar to Shiomi et al.'s life-size teddy bear robot \cite{shiomi2017hug, shiomi2017robot}. HugBot also sits on the floor and requires adults to crawl to it to receive a hug; children can simply bend down. The robot has pressure sensors (four on the chest and two on each arm) to record how much pressure is exerted and an inner t-shaped wooden structure upon which the two soft robotic arms are attached. This sensors and actuators are situated inside a large stuffed animal. These researchers found that zoomorphic robots kept children more engaged compared to non-zoomorphic robots.

Another team of researchers created a wheeled inverted pendulum robot named Robovie-III that hugs in a multi-step process \cite{miyashita2004human}. It uses a range sensor to determine the distance between itself and the user. Once it is the appropriate distance away, the robot hugs the user. After some time, the robot releases. The robot uses the human to assist its balance, and it does not have any padding or softening on top of its metallic components. 

Yamane et al. have been issued a patent for Disney Enterprises, Inc.\ to create their own version of a huggable robot \citep{Disney}. This robot is designed for human interaction, presumably within theme parks. It features a rigid inner structure with specific elements made of softer material to create a deformable exterior in areas that would contact a human. This robot attempts to match the pressure an external user applies using pressure sensors. The wording of the patent is ambiguous as to whether this robot will be autonomous or teleoperated. However, as the users expect the hug to come from the robot itself, not another person, we identify it as human-robot interaction, rather than a technology-mediated hug. The physical appearance of this robot matches that of the character Baymax from the Disney movie ``Big Hero 6'' (2014). This patent supports the idea that there is great interest in furthering human-robot interaction to include more pleasant physical exchanges, particularly human-robot hugging.

\citet{Kaplish2019} created a humanoid robot for physical human-robot interaction. It has two Franka Emika robotic arms fitted with 3D printed shells. These shells are covered in polyurethane foam for a soft and enjoyable tactile experience for the user, and bellows cover the gaps in the elbows and neck. The shells also have embedded optics-based force sensors, which the authors use to estimate the hug tightness. The robot's upper body is finally covered in custom stretchable fabric. In their study, a human operator hugs a mannequin while wearing a suit covered in eight IMUs and the same force sensors that are found on the robot. The robot then hugs another mannequin while being controlled by the sensor measurements pre-recorded from the human operator's suit. In this experiment, the robot never hugged another person and did not hug in real time. The main focus was instead on testing how the sensor readings are mapped from the human's suit to the robot's force and motion control.

Later, \citet{Campbell2020} built on \citet{Kaplish2019}'s work using the same robotic platform, this time testing with users directly hugging the robot. After equipping the robot with 61 force sensors, they trained a sparse learning-from-demonstration model with teleoperated data collected from 121 sample hugs from four participants. They found that their model generalized well to unseen hug styles and new interactions with six human hugging partners (four from training and two new individuals). Interestingly, the authors created solutions to edge cases when the human or robot hugging partner did not behave as expected, such as the robot hugging the air, a delay before closing the arms for a hug, or hugging without making contact (``air hug''). However, when users gave the robot an air hug, the robot did not know when to release the user; thus the user became trapped and had to be released by an experimenter.

Block and Kuchenbecker previously created the original HuggieBot \cite{HuggieBot_master, block2019softness, block2018emotionally}, which applied hardware and software upgrades to a Willow Garage Personal Robot 2 (PR2). They used foam and cloth to soften the robot and various heating elements to warm the robot. A stretchable tactile sensor added to the robot's back detected when a user made contact (indicating the start of the hug) and released contact (indicating a desire to end the hug). They tested combinations of soft and warm hugs and tested the pressure the robot should apply and how long the hug should last. Block and Kuchenbecker found that hugging robots should be soft, warm, squeeze tightly, and release immediately once the user indicates they are ready to be released; users were displeased when released either too soon or too late. These findings formed the basis for the invention of HuggieBot 2.0, a custom hugging robot platform that was recently created and validated by \citet{TheSixHugCommandments}. Because we use it in our action-response elicitation study, this robot is fully described in Section~\ref{UserStudyMethods}. 

None of the human-robot hugging devices described above are able to detect, classify, or respond to intra-hug gestures. As mentioned in Section~\ref{subsec:socialtouchpeople}, humans commonly perform intra-hug gestures during prolonged hugs to provide the hugging partner with beneficial deep pressure touch. Our work in this article builds on all of these findings from prior research and is specifically focused on the exchange of intra-hug gestures between a user and an autonomous adult-sized hugging robot, a challenging social-physical interaction that could greatly enrich hugging robots and provide insights relevant to other affective embodied systems.

\subsection{User Experience Evaluation}
\label{subsec:user-experience-evaluation}
A major part of this work is trying to quantify how users feel about particular versions of HuggieBot. Several different methods exist to gauge how participants feel about technologies. At the most basic level of assessment is forced choice, where participants are given two conflicting alternatives and must choose one \cite{dhar2003effect}. When evaluating a technology, such a force-choice question could be ``I would use this technology in my home,'' where a participant's only options are ``yes'' or ``no.'' 

One step up from forced-choice questionnaires are graded-scale questionnaires \cite{morillo2019journey}. An example of such a question type commonly used in human-robot interaction evaluation is a Likert scale, which usually has between five and nine response points \cite{schrum2020four}. However, Likert scales typically do not provide highly granular responses. An increased number of response points increases mental fatigue on participants and can thus lower the overall quality of responses \cite{bendig1953reliability, lee2014search}. An example of a Likert-scale question might be ``I would use this technology in my home,'' where a participant must answer from the following options: ``strongly disagree,'' ``disagree,'' ``neutral,'' ``agree,'' and ``strongly agree.'' 

For our questionnaires, we chose to use a continuous sliding scale, which is also known as a visual analog scale (VAS) \cite{bijur2001reliability}. Each end of the scale was anchored, for example, from hate (0) to love (10), with neighboring responses separated by only 0.1. Thus, compared to Likert scales, we were able to obtain granular responses without increasing the mental fatigue on our users or lowering the quality of their responses \cite{chyung2018evidence, adamchic2012psychometric}. All of our continuous sliding scales defaulted to start at neutral (for example at 5) so as not to sway our users toward either end of the spectrum. Having a continuous sliding scale with a default at neutral also ensures users do not feel pressured into making a choice that does not feel right to them; they are free to leave the slider at neutral. \citet{wall2017reliability} also found that use of a continuous sliding scale had higher inter-rater reliability compared to a seven-point Likert scale. A final benefit of continuous sliding scale questionnaires is that VAS data have ratio scale properties, which support the use of parametric statistical analysis \cite{myles1999pain}.

\citet{lindblom2020evaluating} discuss how to evaluate the user experience in HRI studies. They make important distinctions between evaluating ``what users say'' (attitudinal) and measuring ``what users do'' (behavioral), as well as understanding ``how and why'' (qualitative) and ``how often or how much'' (quantitative). Some assessment can be done through surveys, as discussed above, but they mention that surveys alone cannot fully encompass the user experience. Naturalistic field studies are recommended to understand how users respond to the technology in real-life conditions. \citet{Alenljung_UX} explain that common methods of user evaluation in HRI include scenario-based evaluation, questionnaires, interviews and focus groups, Wizard-of-Oz studies, expert evaluations, and physiological measurements.

Since some of the richest data collected regarding the user experience cannot be collected via surveys, and since both \citet{lindblom2020evaluating} and \citet{Alenljung_UX} mention the advantages of open-ended questions and interviews, we supplement our surveys with both. Users in both of our studies wrote responses to open-ended questions and then called over an experimenter and verbally explained their answers. While we were unable to perform a naturalistic field study given the current global pandemic, our validation study (Section~\ref{Validation}) included two free-play scenarios during which we observed ``what users do'' and ``how often or how much,'' as suggested by \citet{lindblom2020evaluating}.

\citet{Bargas2011} reviewed 51 papers that were published in the human-computer interaction (HCI) literature from 2005 to 2009, reporting 66 studies that were focused on evaluating user experience. They found that many studies use self-developed questionnaires to evaluate user emotions, enjoyment, and opinions of system aesthetics. \citet{Bargas2011} mention that though it is hardly asked, a crucial element of user experience is understanding the context of use and the anticipated deployment of a system. Following the guidelines by \citet{Bargas2011}, we report all interview questions and protocols, evaluate a prolonged interaction (longer than 30 minutes), and examine and report user behavior (what people do) in addition to what they say.

%% file: sections/03_design_guidelines.tex
\section{Design Guidelines and Research Goal}
\label{DesignGuidelines}
To advance the state of the art in robotic hugging, this article proposes and evaluates a refined version of tenet 4 presented by \citet{TheSixHugCommandments} plus five new design guidelines for the future creation of hugging robots. 

\begin{itemize}

\item[G4.] (refined) When a hugging robot is the one initiating the interaction, it should \textit{autonomously invite the user for a hug} when it detects someone in its personal space. A hugging robot should \textit{wait for the user to begin walking toward it} before closing its arms to ensure a consensual and synchronous hugging experience. 

\item[G7.] A good hugging robot should \textit{perceive the user's height and adapt its arm positions accordingly} to comfortably fit around the user at appropriate body locations.  

\item[G8.] It is advantageous for a hugging robot to \textit{accurately detect and classify gestures applied to its torso in real time}, regardless of the user's hand placement. 

\item[G9.] Users like a robot that \textit{responds quickly to their intra-hug gestures}. 

\item[G10.] To avoid appearing too robotic and to help conceal inevitable errors in gesture perception, a hugging robot should not attempt perfect reciprocation of intra-hug gestures. Rather, the robot should adopt a \textit{gesture response paradigm that blends user preferences with slight variety and spontaneity}. 

\item[G11.] To evoke user feelings that the robot is alive and caring, the robot should \textit{occasionally provide unprompted, proactive affective social touch} to the user through intra-hug gestures. 
\end{itemize}

\textbf{Research Goal:} We seek to evaluate the extent to which each of these six new design guidelines can improve user perception of hugging robots.

\textbf{Research Process:} To test these guidelines, we went through a cycle of collecting a large corpus of sensor data from 32 users, updating HuggieBot 2.0's core hardware and software with extensive pilot testing to address user comments and experimenter observations, creating an intra-hug gesture detection and classification algorithm, developing a probabilistic behavior algorithm for responding to intra-hug gestures, and testing both algorithms on HuggieBot 3.0 in real time with sixteen new users. 

All human-robot interactions exist somewhere along a spectrum from being highly unnatural and protocol-centric to being fully natural and socially intelligent. As discussed in Section~\ref{subsec:user-experience-evaluation}, different experimental approaches from the literature fall in different positions on this spectrum. The work presented in this manuscript is not completely protocol-free. However, with each iteration of HuggieBot (two of which are discussed in this paper), we are moving the interaction closer toward the natural end of the spectrum.  Specifically, the overall experience of hugging HuggieBot 3.0 is more natural than any of the solutions described in the existing literature, as reviewed in Section~\ref{subsec:human-robot-social-touch}. We hope that these validated guidelines and our discussion of detailed quantitative and qualitative findings can help roboticists design human-robot hug interactions that are more natural and human-like than the current state of the art.

%% file: sections/04_materials_and_methods.tex
\section{Action-Response Elicitation Study -- Methods}

This study serves three main goals.  First, we seek additional user comments on all aspects of HuggieBot 2.0 to guide major updates to this platform. Second, this study aims to collect a large corpus of representative haptic sensor data for four common intra-hug gestures so that we can create a perceptual pipeline that detects and identifies these gestures in real time. Third, this study seeks to gather user preferences for how a hugging robot should respond to intra-hug gestures; these opinions will be used to develop a behavior algorithm that enables autonomous hugging robots to respond well to intra-hug gestures that the user performs.

\label{UserStudyMethods}
\subsection{Robot Platform} 

\begin{figure}[t]
\includegraphics[width=0.9\columnwidth, trim = {0cm 0cm 0cm 0cm}, clip]{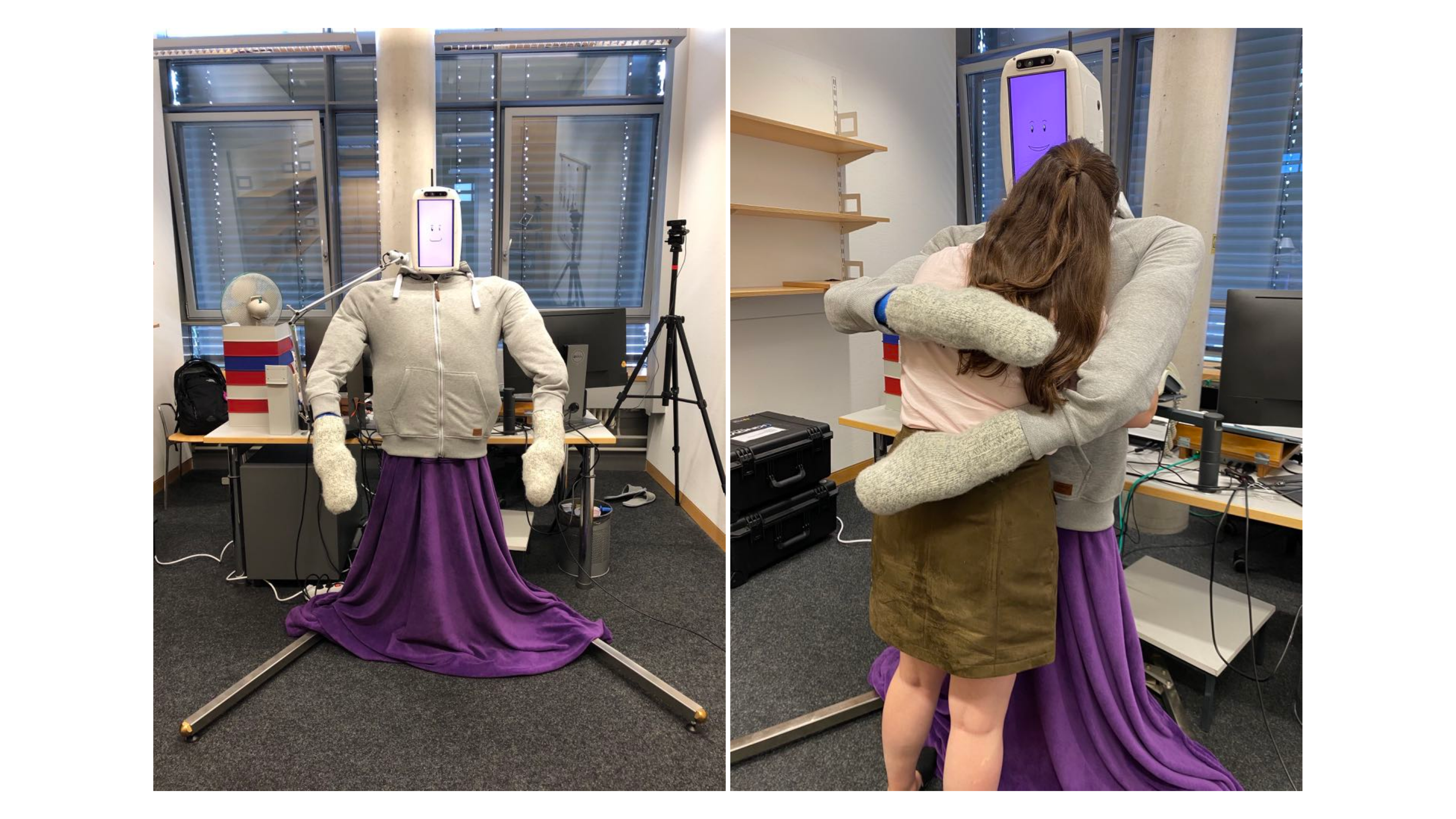}
\vspace{-0.3cm}
\caption{Views of HuggieBot 2.0~\cite{TheSixHugCommandments} ready for a hug and hugging a user. This custom human-sized hugging robot has two padded arms, an inflated torso, and a face screen mounted to a rigid frame. A camera above the screen visually senses the user at the start of the interaction, and torque sensors on the shoulder flexion and elbow flexion joints are used to embrace the user with a comfortable pressure. A microphone and pressure sensor in the back chamber of the torso are used to detect user contact and detect and classify gestures. The user ends the hug by releasing the robot's torso and/or leaning back against the arms.}
\label{fig:UserHug}
\vspace{-0.3cm}
\end{figure}

\begin{figure}[t]
\includegraphics[width=0.8\columnwidth, trim = {0cm 0cm 0cm 0cm}, clip]{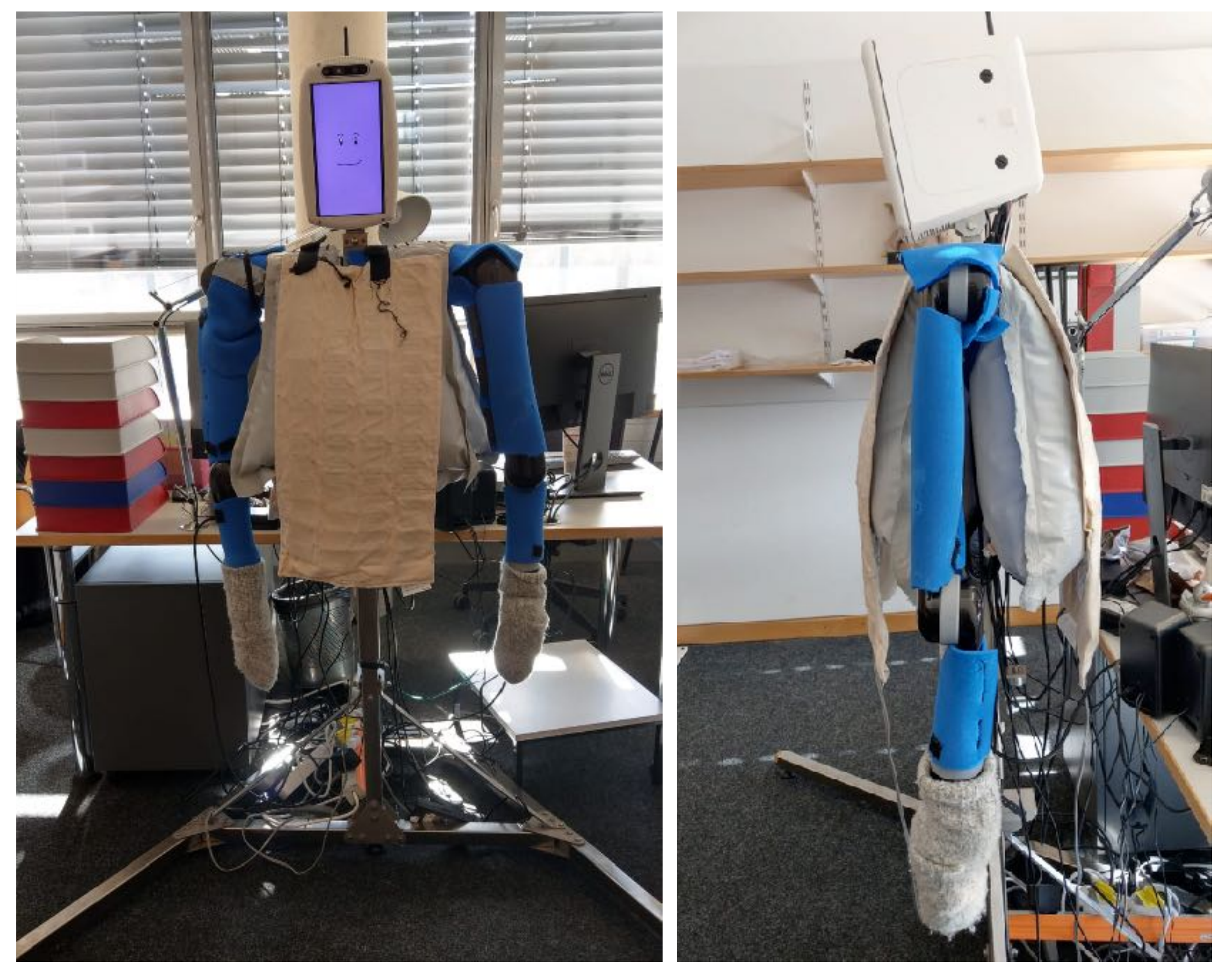}
\vspace{-0.3cm}
\caption{Front and side views of HuggieBot 2.0~\cite{TheSixHugCommandments} without the robe and sweatshirt so that the HuggieChest, heating pads, and foam arm padding can be clearly seen.}
\label{fig:NakedRobot}
\vspace{-0.3cm}
\end{figure}

Block et al.\ previously created and validated a custom hugging robot called HuggieBot 2.0 \cite{TheSixHugCommandments}, which we use in this study. As shown in Fig.~\ref{fig:UserHug}, the robot features a v-shaped base that makes it easy for users to get close for a hug. The robot's torso, HuggieChest, is a custom sensing system that simultaneously softens the robot and detects user contacts. It is made of two chambers of sheet polyvinyl chloride (PVC) that are fabricated using a combination of heat sealing and gluing. Inside the chamber located on the back (where users place their hands) is an Adafruit BME680 barometric pressure sensor and an Adafruit electret microphone amplifier MAX4466 with adjustable gain. Both are connected to an Arduino Mega, which sends the data to ROS (Robot Operating System) over serial at about 45~Hz. The pressure sensor is used to detect the start and end of user contact with the back chamber, and both sensors will be used to detect and classify intra-hug gestures in this article. As explained by \citet{TheSixHugCommandments}, the combination of a microphone and pressure sensor was selected to create a simple and soft omnidirectional haptic sensor. Since users come in different shapes and sizes and have different hugging preferences, it is important that the system can detect contact over a large area, regardless of where the user touches the surface of the robot's torso. In this case, contact detection was more important than contact localization, though localization would be an interesting aspect to study in the future. We also opted for a simpler sensing system to minimize the transmitted data, conserve computational power for real-time processing, and facilitate replication.

The robot has two six-degree-of-freedom Kinova JACO arms mounted horizontally on a custom metal frame. These arms were selected for being anthropomorphic, quiet, and safe; their movement speed is limited in firmware, and they can be mechanically overpowered by the user if necessary. Torque sensors at each arm joint are used to automatically adjust to the user's size at the start of the embrace, and torque signals above a threshold give the user a second way to end the hug. As shown in Fig.~\ref{fig:NakedRobot}, the arms are covered in foam pads for softening. The robot's head is a custom 3D-printed case that houses the Dell OptiPlex 7050 minicomputer that controls the entire robot, the Intel RealSense Depth Sensing Camera, the robot's face screen, a small JBL speaker, and many cables. On top of each torso chamber is a Thermophore MaxHeat Deep-Heat Therapy heating pad (35.6 cm $\times$ 68.6 cm). A purple robe and gray sweatshirt are placed on top of the heating pads, and gray mittens cover the robot's end-effectors to create the final robot outfit. Further details of HuggieBot 2.0's hardware design can be found in \citet{TheSixHugCommandments}.

The minicomputer runs Ubuntu 14.04 and controls the robot via ROS Kinetic. All six robot joints on each arm have angle and torque sensors, which we continuously monitor and record. We use a PID controller to command each joint angle over time. We selected a set of human-inspired target angle waypoints for each arm to move through to create the robot's hugging motion. The robot arms start at its side while the camera module waits to detect a user. Once detected, the robot asks the user for a hug saying, ``Can I have a hug, please?'' while the robot's face changes to show the mouth animation. Then, the hug is executed by commanding each joint to move at a fixed angular velocity toward a predetermined goal pose through the set of target angle waypoints. To adjust haptically to the size of each user, the robot arms move toward a pose sized for the smallest anticipated user. We continually monitor each joint torque and stop a joint's movement if the joint torque exceeds the pre-set torque threshold. When the torque is exceeded, we give each joint a new target angle, which is the joint angle where the torque was exceeded. This method haptically adjusts the robot's embrace to the size of each user (T5 from \citet{TheSixHugCommandments}). 

\subsection{Ethical Approval, Recruitment, and Participants}
The Ethics Council of the Max Planck Society approved this study under protocol F006B of the Haptic Intelligence Department's framework agreement. The first author recruited participants by email, social media, and paper flyers. All participants were English-speaking volunteers from the local area in Stuttgart, Germany. We ran two subjects as pilot participants to refine the experimental methods; their data were excluded from analysis because they were not given the same instructions as the later participants. 32 people participated in the study; twelve males and 20 females. The participants ranged in age from 21 to 60 (mean = 30, standard deviation = 7) and came from thirteen different countries. Overall, the participants did not have a technical background; many experienced their first interaction with a robot during this study. The 27 participants not employed by the Max Planck Society were compensated at a rate of 8 euros per hour.

\subsection{Procedure}
After confirming their eligibility to participate given the exclusion criteria and local COVID-19 regulations, users scheduled an appointment for a 1.5-hour-long session with HuggieBot 2.0. After the participant's arrival, the experimenter explained the protocol using the informed consent document as a guide. The potential subject was given time to read over the consent form thoroughly and ask any questions. If they were still willing to participate, the subject signed the consent form and the video release form. After receiving both these documents, the experimenter turned on two video cameras to record the experiment. 

The user filled out a demographic survey on a tablet computer. Then, the investigator introduced the robot as the personality ``HuggieBot,'' told users it was ``a robot that loves to hug,'' and explained its key features, including the emergency stop. The recruitment materials coupled with this introduction helped set the expectation that the robot's hugging behavior was considered normal and emotionally positive, allowing us to achieve affective grounding with our participants \cite{jung2017affective}.  The experimenter explained how the first half of the experiment would work and how the participant could initiate the hug. The experimenter also explained that the user is always in control of the duration of the hug and explained the different ways to non-verbally cue the robot that they wanted to end a hug (release hands from the robot's back or lean back against the robot's arms). At this point, the subject filled out an opening survey to document their initial impressions of the robot. Users answered the same questions at the end of this experiment; these opening and closing survey results were previously published by \citet{TheSixHugCommandments}, showing significant positive changes in several ratings. Users practiced hugging the robot and acclimated to the hug initiation methods and the timing of the robot's arm movements before the experiment began. All users then experienced eight hugs that made up the first half of the experiment; these results were also previously published, validating the haptic sensing for hug sizing and hug release \cite{TheSixHugCommandments}. Interestingly, Block et al.'s analysis of these results found that users showed no preference between starting the hug with a button press or starting it by walking toward the robot \cite{TheSixHugCommandments}. We seek to improve the latter hug initiation method in this article, as will be discussed in Section~\ref{Improvements}.

The second half of the experiment contained the activities related to intra-hug gestures. We used two 4$\times$4 balanced Latin squares and a participant number that is a multiple of eight to counter-balance any effects of presentation order \cite{LatinSquare}. Participants experienced a total of sixteen hugs in four groups, each made of four hugs. In each group, the user was instructed to perform a specific action (hold still, rub the robot's back, pat the robot's back, or squeeze the robot) at any point during the hug, as many times as they desired. The experimenter provided the same narrative to each user, to ``perform the gesture like they would on a friend or a family member during a hug.'' Within a group of hugs, in response to a user action, the robot would perform a different gesture during each hug (staying still, moving vertically, tapping on the user's back, or tightening its hold on the user). When staying still, the robot's arms did not move. For moving vertically (rubbing), the shoulder lift angle of the robot's left arm was increased by 3$^\circ$ and then returned to its original value twice in a row for each rub response.  For tapping on the user's back (patting), the elbow flexion joint of the robot's left arm was increased by 3$^\circ$ and then decreased by 6$^\circ$ twice in a row before returning to its original value.  For tightening the hold on the user (squeezing), the shoulder flexion joints of both arms were adjusted inward by 1$^\circ$ while both elbow flexion joints were simultaneously adjusted inward by 5$^\circ$; all four joints were then returned to their original values. Each movement is commanded in joint space relative to the current hug's embracing pose around the user. The joints move between points at the robot's maximum speed, yielding fixed-duration gestures that last approximately 2 seconds. 

Because we had not yet developed autonomous intra-hug perception or action capabilities, the timing of each robot response was controlled by the experimenter, who visually observed the actions of the user. A version of Fig.~\ref{fig:teaser} was printed on a large poster and placed in the experiment room for the participants to reference at any time. While the robot responses were designed to be the same as what the user was instructed to perform, the poster contained only the pictures and descriptions, without the colored gesture names, to avoid swaying the participants to match robot responses with user actions of the same name. After each hug, the experimenter asked the user which intra-hug gesture they thought the robot had performed; the hug was repeated if the user was not able to identify the robot response correctly, or if the user performed the wrong action.

After a successful action-response hug, users were asked to rate how much they liked that robot response, given the action they performed, using a continuous sliding scale from hate (0) to love (10) with a resolution of 0.1. To focus on more general principles of human-robot interaction, here we asked users to rate the appropriateness of the gesture response rather than the quality with which HuggieBot 2.0 performed the gesture.  After users had experienced all four robot responses for the given user action, they were given time to review and adjust their survey entries before calling the experimenter to verbally explain their ratings. After testing all sixteen hug combinations of user action and robot response, participants were asked to rate the quality of each of the robot responses on the same hate-love scale. After the closing survey, a free-response question asked the user to provide any comments or feedback they had about the experiment.

%% file: sections/05_results.tex
\section{Action-Response Elicitation Study -- Results}
\label{Results}
We analyzed the user ratings, pressure sensor and microphone data, and user comments from the action-response elicitation study to understand how HuggieBot 2.0 might be upgraded to become capable of autonomously detecting and responding to intra-hug gestures.

\subsection{User Ratings}
\begin{figure}[t]
\includegraphics[width=0.865\columnwidth, trim = {0cm 0cm 0cm 0cm}, clip]{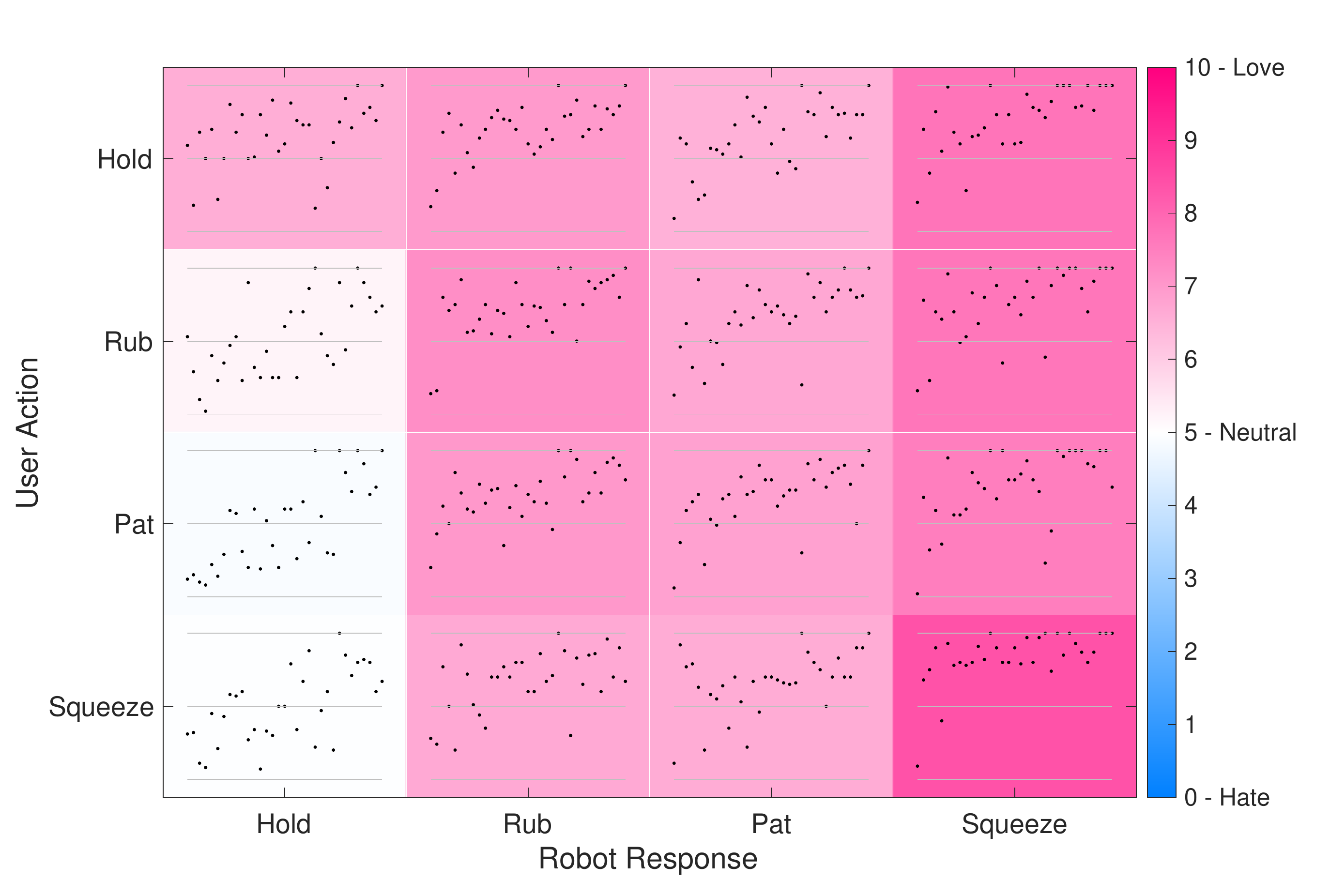}
\vspace{-0.3cm}
\caption{A matrix showing the user ratings of the appropriateness of each possible robot response to the four intra-hug actions that users performed in the action-response elicitation study. The color of each square represents the average rating, following the legend shown at right. The dots in each cell show the individual rating of each user, consistently ordered based on their average score from low to high. The pale horizontal lines in each square show the ratings of 0 (hate), 5 (neutral), and 10 (love).}
\label{fig:BehaviorMatrix}
\vspace{-0.3cm}
\end{figure}

\begin{figure}[t]
\hspace{0.9cm}\includegraphics[width=0.8\columnwidth, trim = {5cm 15cm 0cm 15cm},clip]{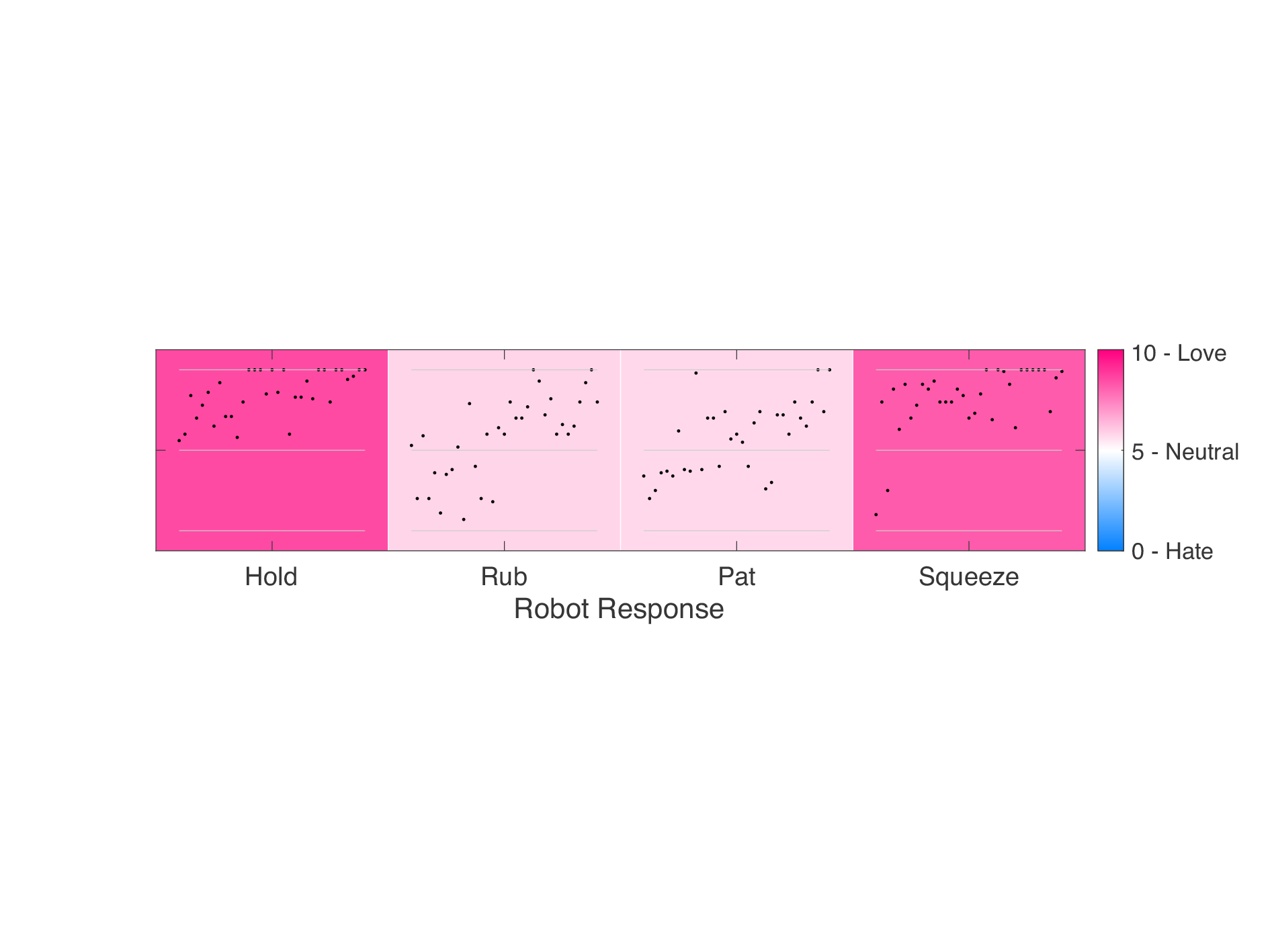}
\vspace{-0.3cm}
\caption{A matrix showing the user ratings of the quality of the four robot responses, using the same visualization approach as Fig.~\ref{fig:BehaviorMatrix}.}
\label{fig:QualityMatrix}
\vspace{-0.3cm}
\end{figure}

As previously mentioned, each user performed each action during four different hugs, experiencing a different robot response during each hug. In total, all users rated sixteen pairs of user actions and robot responses. Figure \ref{fig:BehaviorMatrix} shows the responses to the sixteen different pairs. The color of each cell in the matrix represents the average score from hate (0) to love (10) over all users. The black dots inside each cell show the 32 individual user ratings, always presented in the same order from lowest to highest average user rating. The three lowest average ratings (5.2, 4.9, and 4.9) all occurred when the user performed an active gesture (rub, pat, or squeeze, respectively) and the robot did not move in response (hold). The hold-hold pairing received a much higher average rating (6.6), as did all conditions wherein the robot responded to user inaction (hold) or action (rub, pat, squeeze) with a rub, pat, and especially a squeeze. While some users gave each action-response pair a rating below neutral, the average ratings achieved for the appropriateness of all of the responsive robotic intra-hug gestures were consistently high. 

Figure \ref{fig:QualityMatrix} shows the user ratings of the quality of the robot gestures. Users rated the quality of the robot staying still and squeezing as very high (8.5 and 8.2, respectively); no user gave a negative rating for the robot's hold, and only two of the 32 users gave negative ratings for squeeze.  Rubbing and patting were rated positively (5.8 and 5.7, respectively), but closer to neutral. The fact that these gesture quality ratings differ somewhat from the average response appropriateness ratings in Fig.~\ref{fig:BehaviorMatrix} shows that users were at least moderately successful at distinguishing these rating tasks from one another, particularly regarding hold.

\begin{figure}[t]
\includegraphics[width=\columnwidth, trim = {1cm 4cm 1.5cm 4cm},clip]{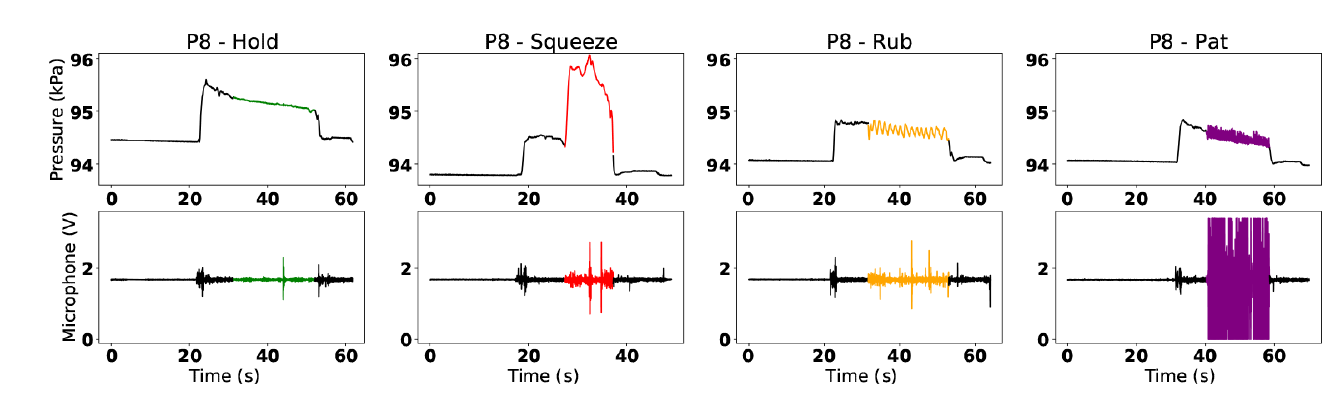}
\vspace{-0.3cm}
\caption{Sample pressure signals and microphone signals from participant 8 performing each of the four gestures during the action-response elicitation study. The colored data points mark the time periods manually labeled as positive examples of the indicated intra-hug gestures.}
\label{fig:P8}
\vspace{-0.3cm}
\end{figure}

\begin{figure}[t]
\includegraphics[width=\columnwidth, trim = {1cm 4cm 1.5cm 4cm},clip]{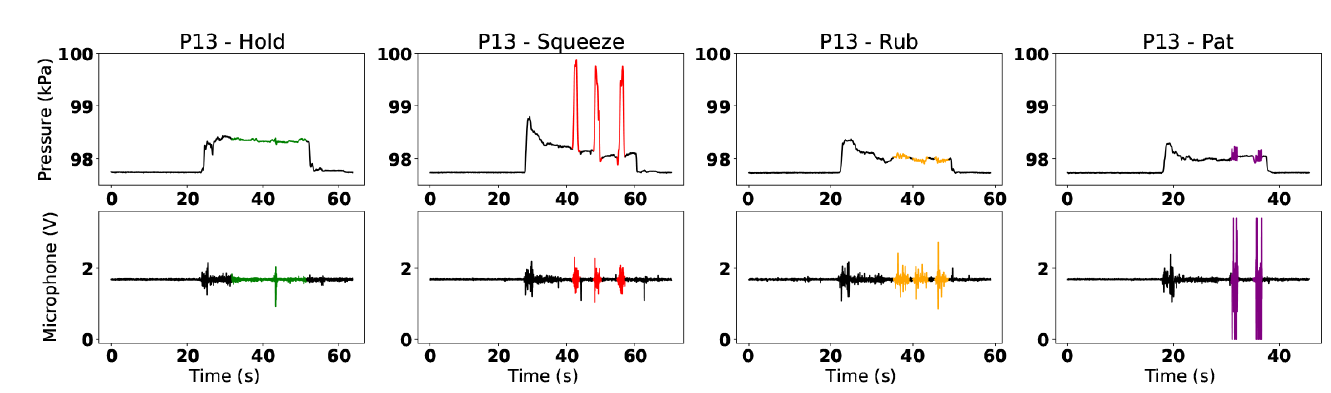}
\vspace{-0.3cm}
\caption{Sample pressure signals and microphone signals from participant 13 performing each of the four gestures during the action-response elicitation study. The colored data points mark the labeled data segments.}
\label{fig:P13}
\vspace{-0.3cm}
\end{figure}

\subsection{Pressure Sensor and Microphone Data}
Data were recorded from HuggieChest's pressure sensor and microphone for each of the sixteen hugs for all 32 users, yielding a total of 512 recordings. Analyzing these signals after the experiment shows common characteristics between different users performing the same gesture. As a representative sample, Figures \ref{fig:P8} and \ref{fig:P13} show the pressure and microphone data collected from two different participants performing the same gestures. While the torso chamber was inflated to somewhat different initial inflation levels for each participant, characteristic signals can still be recognized to determine the type of contact made, regardless of inflation. The use of these two sensors allows us to differentiate between both coarse and fine contacts. While the pressure signals look similar between rubs and pats, we can differentiate between the two by the pat's much larger microphone response. Additionally, the microphone signals look quite similar for squeezes and rubs, but looking at the corresponding pressure signals allows us to determine which action is being performed; squeezing the robot drastically increases the chamber's pressure, while rubbing it does not. These results show the benefit of using two sensors for detecting intra-hug gestures. 

Video review revealed that these two participants had different approaches to performing the gestures, which can be recognized in the recorded data. P8 (Fig. \ref{fig:P8}) squeezed the robot only once and performed a continuous strong patting movement. In contrast, P13 (Fig. \ref{fig:P13}) performed three distinct squeezes and two pats, separating each repeated gesture with a short pause (hold). The variety of ways in which the 32 users performed these four gestures was surprising and underscored the importance of gathering a large corpus of sensor data. 

Finally, although we asked users to perform only a single type of gesture per hug, we noticed that seven of the 32 participants (21.9\%) accidentally combined gestures and sometimes performed two gestures at once. This particularly happened during rubs and pats, when the user would sometimes also unintentionally squeeze the robot.

\subsection{User Comments}
\label{subsec:user-comments-study1}
Our users' written and spoken comments provide crucial information on how to improve the quality of the hugs HuggieBot can deliver. A systematic analysis reveals several key themes repeated by many users. The majority of users (68.75\%) commented that they preferred \textit{not} having to press a button to initiate the hug. The slow speed of the robot's arms and thus the amount of time that it took for the arms to close around the user detracted from the experience (34.37\%), particularly for hugs initiated when the user walked toward the robot. Next, almost half of the users (43.75\%) mentioned that they could not feel both arms fully against their backs in at least one hug. Some users (21.87\%) also commented that the robot's hand placement was inappropriate -- either too high on the back (close to the neck) or too low (on the buttocks), both of which made them uncomfortable. Incomplete arm closure and too high or too low hand placement made it difficult for some users (21.87\%) to feel the robot performing gestures on their back, which most likely contributed to the variety of user ratings reported. These comments show that improvements are needed for how HuggieBot initiates hugs and adapts to its hugging partner.

Participants in this study experienced the robot both responding to their gestures and holding still (not responding) when they performed a gesture. Almost all users (78.13\%) commented that having the robot respond to their gestures made it feel more ``alive,'' ``social,'' and/or ``realistic''. As we initially expected users to prefer a robot that reciprocates their gestures, we were surprised to find users enjoyed variety in the responses. When they explained their response ratings to the experimenter, twenty of our users (62.5\%) mentioned that reciprocation of their actions every time felt ``too mechanical.'' Rather than thinking the robot made a mistake when it performed a different gesture, users appreciated gestures of similar ``emotional investment levels'' (P21) and felt it showed ``the robot understands [them] and makes his own decision'' (P30). 

In agreement with the high ratings shown in Fig.~\ref{fig:BehaviorMatrix}, many users shared positive comments about being squeezed by the robot, saying phrases such as ``I love warm, tight hugs'' (P7, P8, P17, P21, P25) or that being squeezed by the robot felt ``the most natural'' (P15) and ``the closest to a real human hug, the best response'' (P2, P16, P20). Some even went as far as to say that the squeezes gave them ``a sense of security and comfort'' (P9, P20, P23). In general, users thought the duration of the robot's timed squeeze was ``too short'' (P5, P24). Additionally, several users suggested making the robot's squeeze duration match theirs because the fixed timing felt ``too mechanical'' (P1, P6) or like ``the robot wasn't as emotionally invested as [they] were'' (P21, P28). These comments hint at the need to treat modal gestures such as squeeze differently from event-based gestures such as rub or pat.    

Finally, another unexpected finding that wove throughout the comments was how strongly the users anthropomorphized the robot. The experimenter always called the robot ``it'' or ``HuggieBot,'' yet in both written and verbal comments, 96.87\% of users referred to the robot as a ``he'' when describing how the hugs felt, often explaining a social situation it reminded them of. Such interactions included ``a comforting hug from a mother'' or ``a distant relative at a funeral,'' ``seeing friends at a football match,'' ``receiving a pity hug from someone who doesn't want to,'' ``hugging an ex,'' to even ``hugging a lover.'' They attributed emotions, mood swings, and attitudes to the robot, depending on how well it hugged them in each trial of the study. 

\subsection{Brief Discussion}
The ratings gathered in this study provided essential information about user preferences for how a hugging robot should perform intra-hug gestures. We conducted this experiment to guide hardware and software improvements to HuggieBot 2.0, as will be discussed in the following Sections. However, the results can also be generalized and applied to other hugging robots, including those that do not have any haptic sensing. As can be seen in Fig.~\ref{fig:BehaviorMatrix}, regardless of the user action, a squeeze was always perceived on average as the most enjoyable robot response, including when the user had not just actively performed an action (after user hold). Therefore, researchers who want to improve user opinions of their hugging robot without investing in haptic sensing capabilities should program their robot to occasionally squeeze the user. Simultaneously, the neutral average reactions users showed when the robot did not respond to their rubs, pats, and squeezes indicate that perceiving and responding to intra-hug gestures could greatly improve hugging robots. 

%% file: sections/table_huggiebot.tex
\begin{table}[tb]
\caption{Summary of HuggieBot 2.0 and 3.0 features.}
\label{tab:huggiebot}
\centering
\footnotesize{
\begin{tabular}{p{0.1\textwidth}p{0.41\textwidth}p{0.41\textwidth}}
\hline
&\textbf{HuggieBot 2.0~\cite{TheSixHugCommandments}}&\textbf{HuggieBot 3.0 (new)}\\
\hline
\rule{0pt}{3ex}\textbf{Hug}&Single-stage initiation: the experimenter tells the&Consensual two-stage initiation: After detecting  \\
\textbf{initiation}&user when the robot is ready. HuggieBot initiates a hug either when the user presses a key or when it visually detects that a user is walking toward it. The robot speaks after the hug has been initiated. The user must adjust their walking speed to match HuggieBot's movements and wait for the arms to close.&a user in its personal space, HuggieBot opens its arms and verbally invites the user for a hug. HuggieBot waits for the user to start walking toward the robot before starting to close its arms.\\

\rule{0pt}{3ex}\textbf{Arm}&Constant arm lift angles suitable for an adult of&Automatic adaptation of arm lift angles to the user's \\
\textbf{placement}&average height.&observed height (equations \eqref{eq:1} and \eqref{eq:2}).\\

\rule{0pt}{3ex}\textbf{Torso}&Two warm inflatable chambers with a smaller chamber in the back compared to the front. Each chamber has a microphone and a pressure sensor, with wires exiting at the bottoms of the chambers.&Two warm inflatable airtight chambers of equal size on the front and back. Only the back chamber has a microphone and a pressure sensor, and the wires exit from the top of the chamber.\\


\rule{0pt}{3ex}\textbf{Embrace}&Automatic adaptation of arm closure to the user's size by thresholding individual joint torque measurements on the four involved arm joints. Wrist flexion joints oriented horizontally, which causes local high pressure on the user's back at the points of contact with the wrist.&Automatic adaptation of arm closure to the user's size with better robustness and comfort through filtering of torque values, adjusted torque thresholds, and adjusted final arm closure angles for the four involved arm joints. Wrist flexion joints oriented vertically so that the surfaces in contact with the user's back are flat.\\
  
\rule{0pt}{3ex}\textbf{Hug}&Dual activation: the user removes their hands&Dual activation: same as HuggieBot 2.0 with a\\ \textbf{release}&from the robot's back (pressure threshold) and/or leans back against the robot's arms (torque threshold).&reduced torque threshold to allow for easier release.\\

\rule{0pt}{3ex}\textbf{Gesture detection}&Wizard of Oz (by a human operator)&Autonomous: random forest algorithm based on one second of data from the microphone and pressure sensors (Section~\ref{Detection}).\\

\rule{0pt}{3ex}\textbf{Classified gestures}&Wizard of Oz (by a human operator)&Autonomous: random forest algorithm classifies each time window into hold, rub, pat, squeeze, squeeze-pat, or squeeze-rub (Section~\ref{Detection}).\\

\rule{0pt}{3ex}\textbf{Behavioral response}&Wizard of Oz (by a human operator)&Autonomous: probabilistic response based on mean user preference ratings with slight variety (Fig.~\ref{fig:BehaviorMatrix}, equation~\eqref{eq:3}), including some pro-active gestures during periods when no user gestures are detected (Section~\ref{Response}).\\

\hline
\end{tabular}
}
\vspace{-.15in}
\end{table}

%% file: sections/06_changes_to_platform.tex
\section{Improvements to the Platform}
\label{Improvements}
The results of the previously published validation of HuggieBot 2.0 \cite{TheSixHugCommandments} and the user comments from the action-response elicitation study (Section ~\ref{subsec:user-comments-study1}) showed four main aspects of the system that could benefit from improvement: the hug initiation process, the vertical placement of the robot's arms on the user's body, the reliability of the inflated torso, and the quality and consistency of the robot's embrace around the user's body. Thus, we spent time addressing these concerns to improve the quality of the hug that this robot can deliver to users, upgrading HuggieBot from version 2.0 to version 3.0. Table \ref{tab:huggiebot} summarizes the key features of these two successive versions. We extensively piloted all of these changes with representative users and made further adjustments based on their feedback. The following subsections provide more detail about the final changes made and used in the validation study.

\subsection{Hug Initiation}
\label{subsec:initiation}
The initial evaluation of HuggieBot 2.0 by \citet{TheSixHugCommandments} tested two different ways for users to initiate a hug, always starting about 2.5 meters in front of the robot. In the first method, the user pushed a button to start the hugging process. The second method used HuggieBot's built-in depth camera to recognize a human and then start the hugging process when the potential user starts walking toward the robot. Users did not rate the two methods significantly different; this indifference can be explained by their comments on this topic. Because the visual hug-initiation method was triggered by the user's forward movement, and because the robot arms close slowly, users would often reach the robot before its arms had closed very far, causing them to have to wait for the robot's embrace. 

After piloting several alternative approaches, we improved the visual hug initiation process by dividing it into two steps. First, when all the necessary software nodes are running, the robot lifts its arms and asks the user ``Can I have a hug, please?'' The phrase and arm movement clearly show the user when they may begin hugging the robot; previously, the experimenter prompted users when they could begin. Lifting the arms also beneficially reduces the distance the robot's arms need to travel to close around the person, thus shortening the waiting time disliked by users. After this invitation step, HuggieBot uses the previous method to visually detect the user's forward motion and initiate the closing sequence. The robot waits between the ``hug request'' pose and closing its arms for as long as the user needs in order to reduce time pressure, so users do not feel like they have to start the next hug immediately. These small changes were implemented to make the robot's hug timing more natural and intuitive. This method also beneficially reduces experimenter interaction with the user and better mimics human-human hugs, where one person lifts their arms for a hug and waits for their partner to approach before wrapping their arms around them. 

\subsection{Adjustment to User Height}
\label{subsec:height}
As shown in Fig.~\ref{fig:QualityMatrix}, the robot's rubs and pats received lower average quality ratings than hold and squeeze. When users explained the low ratings, the most common criticism was the location at which the gesture was performed, which was not optimal for their bodies. As HuggieBot 2.0's arms always hugged at the same height off the ground, these gestures were performed too low for tall users and too high for short users. In addition to the inappropriateness of some contact locations reported in the comments, the convex or concave curvature of different areas of different users' backs exacerbated this problem by causing loss of contact or excessive contact when the robot performed some gestures.

To resolve this hand placement issue, the robot must improve its visual perception of the user. In addition to detecting a potential user and estimating his/her approach speed toward the robot, HuggieBot needs to perceive the user's approximate height and adjust its arm positions accordingly, something humans do naturally, quickly, and efficiently. 

To simplify this problem, several assumptions were made. First, we assumed that the camera is perfectly parallel to the floor. Second, we assumed that the person approaching is standing perpendicular to the floor. These assumptions help simplify the problem from a three-dimensional problem in point-cloud space to a planar problem. Simplicity is desired in this case to keep the computational load low and allow for real-time adjustments. The problem then becomes one of similar triangles. The depth camera's resolution is 1280 pixels $\times$ 720 pixels, and its focal length is 651.55 pixels. Based on the room's size constraints and the need to keep the camera oriented parallel to the floor, the camera cannot see the user's feet and lower legs when the person is first detected. To accommodate this reduction in the bounding box's height, after the visible height of the user has been calculated in meters, a small adjustment is added based on the user's distance from the camera to account for the height of the unseen portion of their body. The full linear projection can be written as follows, using constants obtained through measurements: 
%
%
\begin{equation}\label{eq:1}
H = \frac{(D \cdot b)}{f} - \alpha \cdot D + h_c
\end{equation}
where $H$ is the user's full height in meters, $D$ is the distance between the user and the camera in meters, $f$ is the depth image focal length in pixels, and $b$ is the height of the bounding box of the detected person in pixels. In addition, $h_c = 1.73$~m, which is the height of the robot's camera above the ground, and $\alpha=0.5518$, which is a geometric constant that is multiplied by the distance between the user and the robot to proportionally account for the occluded height between the bottom of the bounding box and the floor. We found that individual height estimates computed in this way are somewhat noisy, so HuggieBot 3.0 averages five successive measurements.


We set the ideal shoulder lift angle for HuggieBot 3.0's left arm based on the estimated height of the user, and we then offset the right arm shoulder lift angle up by 20 degrees from that point to create good inter-arm spacing. To determine appropriate arm lift angles for users of different heights, we performed brief experiments with two model users at the minimum (1.40 m) and maximum (1.93 m) user heights we anticipate encountering. We manually adjusted both robot arms around each model user to a comfortable height on their back and recorded the corresponding shoulder lift joint angles. We perform linear interpolation to find the ideal left shoulder lift joint angle for the approaching user, as follows:
%
\begin{equation}\label{eq:2}
\theta_{\ell} = \theta_{\ell,\min} + (H - H_{\min}) \cdot \frac{(\theta_{\ell,\max} - \theta_{\ell,\min})}{(H_{\max} - H_{\min})}
\end{equation}
where $\theta_{\ell,\min}$ and $\theta_{\ell,\max}$ are the robot's left shoulder lift angle angle for the minimum-height and maximum-height model users, respectively, $H$ is the user's estimated height in meters, and $H_{\min}$ and $H_{\max}$ are the height of the short and tall model users, respectively. When the user's estimated height is outside the range of the model users, the closer model user's robot arm placement is used.

\subsection{New Torso}
\label{subsec:torso}
We created a new inflated torso for HuggieBot 3.0 to address several shortcomings in the previous design. HuggieBot 2.0's torso contained a pressure sensor and a microphone in both the front chamber and the back chamber. Early testing showed that the front chamber data provided little information beyond the back chamber data, so HuggieBot 2.0 did not use the information from these sensors. Our new torso has sensors only in the back chamber; furthermore, the sensor wires exit the top of the chamber rather than the bottom of the chamber to minimize the distance to the computer in the robot's head. 

The torso's initial design featured two different-sized chambers, with the back chamber being slightly smaller. Study participants of all sizes used various arm positions to hug the robot and perform intra-hug gestures on its back. Therefore, we decided to increase the back chamber's size to be equal to the front chamber to better accommodate all users. 

Some users squeezed the robot much more tightly than we anticipated during the squeezing hugs of the action-response elicitation study, occasionally popping holes in a chamber or forcing a resealable inflation valve to open; both of these failure modes allow air to escape, change the feel of the robot's torso, and require re-inflation. We designed the new chamber to be more robust to withstand these higher pressures. We ensured a robust and airtight seal on HuggieBot 3.0's new torso by heat sealing along the edges and then using HH-66 vinyl cement on top of the heat seal. 

The newly constructed torso was tested by pilot users performing the four studied gestures during hugs with the robot. After we matched the sensitivity of the new microphone to that of the previous one, the sensor recordings exhibited the same general patterns as those shown in Figs.~\ref{fig:P8} and \ref{fig:P13}. The lack of leaks in the new chamber beneficially also stabilized the starting pressure, which makes both the feel of the robot and the measurements of its haptic sensors more consistent over time. 

\subsection{Quality and Consistency of the Embrace}
Finally, several smaller changes were made to how the robot's arms grasp the user based on the feedback from the action-response elicitation study as well as additional pilot users. The changes we made are as follows:
\begin{enumerate}
    \item Some users were thinner than we had anticipated, so some of the robot's joints reached their goal angles without coming into contact with the user's back. Therefore, we made the closing goal angles much smaller, such that no user will fit inside the goal pose, which ensures good contact between both robot arms and the user's back. 
    
    \item To automatically adjust to our users' shapes and sizes, we use a torque threshold to turn off individual joints when they come into contact with the user. If a single torque measurement exceeds this threshold, HuggieBot 2.0 stops that joint from moving further. We found that while this kept users safe, it did not provide them with a consistent feeling of being fully embraced because some spuriously high torque readings were occasionally measured. Therefore, we implemented a moving average filter on all torque measurements with a window of three values. When the average torque in this window surpasses the threshold, the joint stops. After making this adjustment, we also tuned the torque thresholds and the relative offset for each joint's final target with multiple diverse users to ensure HuggieBot 3.0's embrace was comfortable and not painful.
    
    \item With the improved arm placement and more complete closure, we noticed that the wrist rotation of the robot's arms could cause uncomfortably high pressure on the backs of some users. We adjusted the goal pose for the wrist angles for both arms to ensure the flat and comfortable side of the wrist is in contact with the user's back. 
    
    \item With the improved quality of the contact between the arms and the user's body, we found that HuggieBot's squeeze was now too tight for many users. We thus reduced the elbow joint's squeeze movement from a magnitude of 5$^\circ$ to only 3$^\circ$. 
    
    \item We reduced the torque threshold required to release from the hug during gestures. This value was set rather high (60~Nm) for HuggieBot to prevent users from accidentally triggering a torque release when performing a gesture; however, pilot users were occasionally bothered by how hard they had to lean back to make the robot release them in the middle of a gesture. This threshold was tuned through pilot testing to have a final value of 20~Nm during a regular embrace and 40~Nm during robot gestures. It is important to note that we kept a single value for this torque threshold rather than checking a moving average of the measurements in order to release the user without delay. If a user wants to end the hug at any time, HuggieBot 3.0 opens its arms immediately. Note that the user presently cannot trigger a release during a timed squeeze from the robot because this call is blocking; however, the user can trigger a release between timed squeezes and also when the robot is performing a squeeze response that is matched in duration to their squeeze.
\end{enumerate}

\noindent These four categories of hardware and software improvements upgraded HuggieBot 2.0 to become HuggieBot 3.0, as summarized in Table~\ref{tab:huggiebot}.

\section{Detection and Classification of Gestures}
\label{Detection}
To build an autonomous perception system that HuggieBot 3.0 can use to detect and classify intra-hug gestures, we manually labeled 498 trials (14 of the initial 512 trials were discarded due to problems in the data) recorded in the action-response elicitation study. One of the authors visualized the pressure and microphone signals and marked the hug's start to be the first point where the pressure rises from its initial value. Similarly, the hug's end was the first point where the pressure value declined back to its initial value. The same author marked the start and end of touch gestures based on the distinct signatures of the gestures in the pressure and microphone signals. 

Next, we divided the time-series hug data into a large number of segments and extracted statistical features from each segment. We applied a moving window (W) with an overlap size (O) to divide the time-series data for each hug into shorter segments. If a segment’s overlap with the gesture timestamps was above a certain threshold (T), the segment was labeled with that touch gesture. For each segment, we subtracted the baseline pressure and baseline microphone values from the corresponding signals. These baseline values were calculated by taking the median of the first 150 pressure and microphone data points after the start of the hug, which corresponds to about 3.3 seconds of data recorded at 45~Hz. We extracted 80 statistical features for each segment, including sum, minimum, maximum, average, median, standard deviation, variance, number of peaks, interquartile range, and area under the curve for the pressure and microphone signals and their first and second time derivatives. 

\begin{figure}[t]
\includegraphics[width=0.75\columnwidth, trim = {0cm 0cm 0cm 0cm},clip]{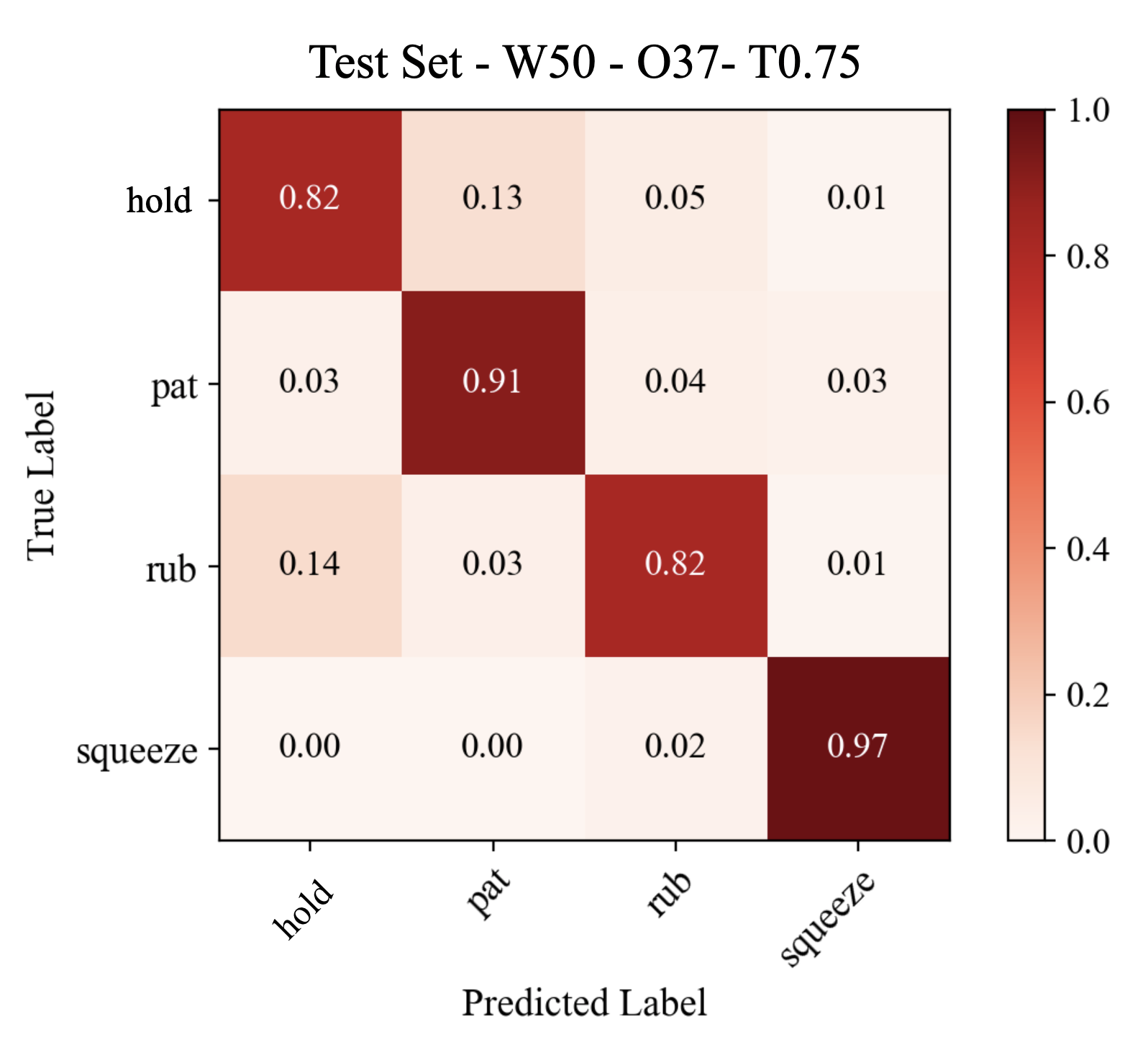}
\vspace{-0.3cm}
\caption{Confusion matrix for the test dataset gathered in the action-response elicitation study.}
\label{fig:confusionmatrix1}
\vspace{-0.3cm}
\end{figure}

We divided the segments into train, validation, and test sets. We trained a random forest algorithm on 70\% of the data and used 20\% of the data as the validation set to determine the impact of the three parameters W, O, and T on the model performance. We tested window sizes of W = 50, 75, and 100 data points, overlap sizes of O = 37, 25, and 12 data points, and labeling thresholds of T = 0.75, 0.5, and 0.25. The classification accuracy varied by only about 1\% to 3\% among different parameter combinations. However, larger values of W and T make the robot slower at detecting user gestures. Thus, we selected W = 50, O = 37, and T = 0.75 
as a trade-off between gesture detection accuracy and delay. Fig.~\ref{fig:confusionmatrix1} shows the confusion matrix after applying the resulting random forest model to the remaining 10\% test data.  The overall classification accuracy achieved is 88\%, with the best performance on detecting squeezes.

After development and offline validation, this perceptual pipeline was transferred to the physical robot and adapted to run in real time.  To conserve computational effort while achieving a fast reaction time, we calculate features on the most recent window of 50 data points every ten samples, rather than every sample. Intra-hug gestures are thus detected at a rate of approximately 4.5~Hz, always yielding an output of hold, pat, rub, or squeeze. After being trained on the dataset collected from HuggieBot 2.0, the classifier was tested on the new inflated torso (Sec.~\ref{subsec:torso}) of HuggieBot 3.0. Pilot testing showed that the classifier's performance was worse than desired, probably due to the changes in the size and shape of the back chamber. Gesture samples were thus acquired from five pilot participants performing each of the four studied gestures during two hugs on the new torso. These 40 trials were labeled and processed in the same way as the original data from the action-response elicitation study. Because these additional examples did not have sufficient power when added to the 70\% training data (354 trials), we trained a new version of the classifier using 80\% of the data from ten randomly selected users from the action-response study (125 trials) plus the newly collected 40 trials. This retrained version showed good performance on the new inflated torso in pilot testing, so it was selected for use in the validation study. 

\section{Behavioral Response to Detected Gestures}
\label{Response}
\begin{figure}[tp]
\includegraphics[width=0.8\columnwidth, trim = {0.1cm 0cm 0.1cm 0.1cm},clip]{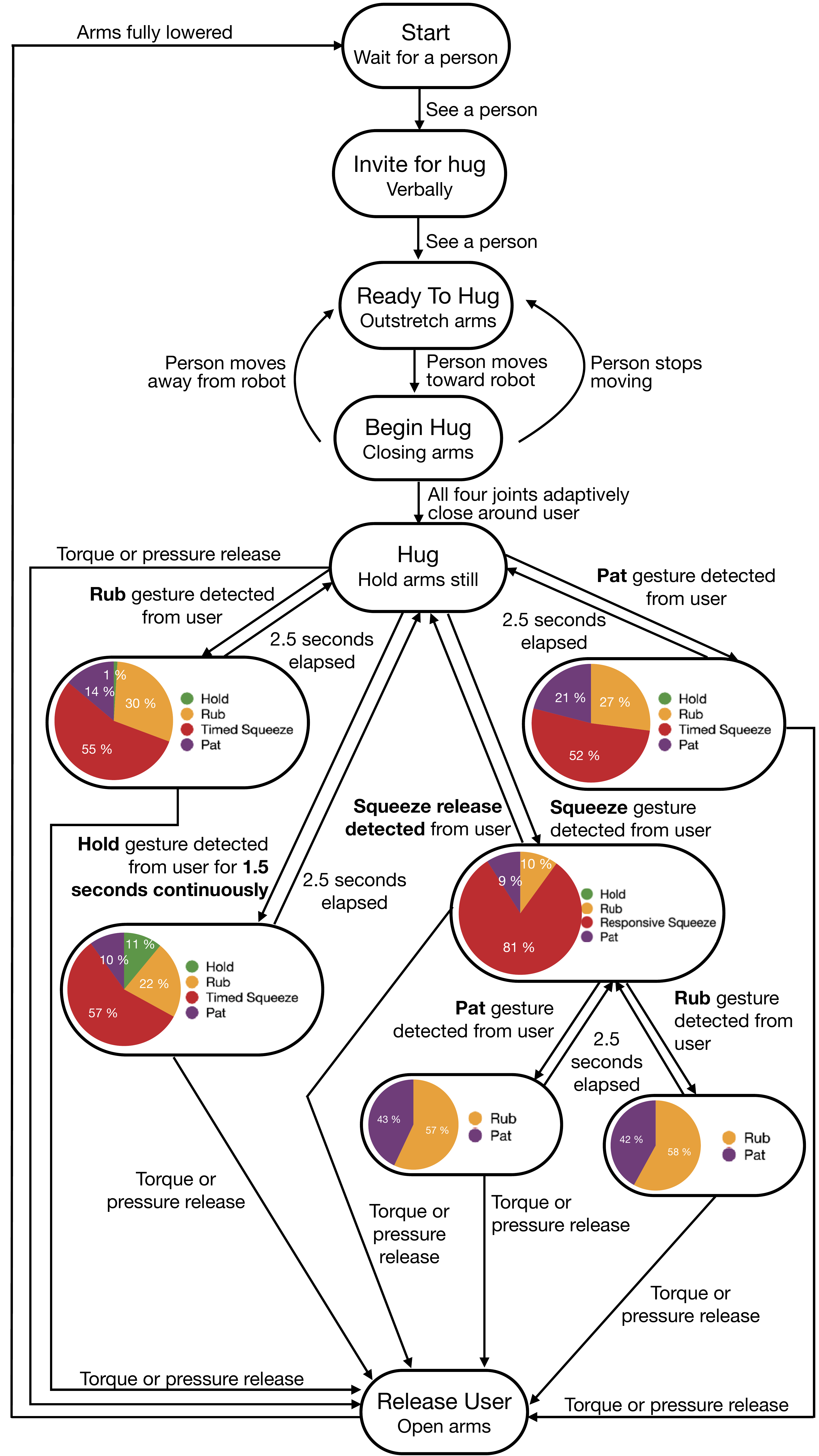}
\vspace{-0.3cm}
\caption{Diagram of HuggieBot 3.0's states and state transitions.}
\label{fig:robot_state_diagram}
\vspace{-0.3cm}
\end{figure}

After it possesses a good pipeline for detecting and classifying intra-hug gestures, a hugging robot needs to decide how to act, i.e., which of the available intra-hug gestures to perform. We implemented the robot's behavior in all situations using a simple probabilistic approach (Section~\ref{subsec:probabilistic}), calculating the likelihood that the robot will perform each gesture as a function of the relevant average user ratings given the action that was just detected. Although we have treated them equivalently thus far, the active gestures of pat, rub, and squeeze occur far less often than the passive gesture of hold in natural hugs. The hold gesture can be thought of as the standard background against which the active gestures occur; almost every hug in the action-response elicitation study included periods of time where the user was simply holding the robot. The differences between active and passive gestures require somewhat different types of behavioral responses from the robot. We found the need for an additional distinction between the discrete active gestures of pat and rub (Section~\ref{subsec:discrete}), which occur in discrete units, and the modal active gesture of squeeze (Section~\ref{subsec:modal}), which involves transitioning into and out of the state of applying higher pressure to the hugging partner. Note that the described methods are not limited to these four intra-hug gestures: another example of a discrete active gestures that future hugging robots could consider is tickling, while an alternative modal active gesture could be leaning into one's hugging partner. Finally, we describe how to apply our probabilistic behavior paradigm when a hugging robot detects a passive gesture such as hold (Section~\ref{subsec:passive}). Figure \ref{fig:robot_state_diagram} depicts the robot's state transition diagram to clarify the robot's controller.

%

\subsection{Probabilistic Behavior Paradigm}
\label{subsec:probabilistic}
Examination of Fig.~\ref{fig:BehaviorMatrix} shows that each row follows a somewhat different distribution of average user ratings; thus, the appropriateness of a particular robot response critically depends on the action the user has just performed. Our probabilistic approach was designed to respect that dependency as well as the relative appropriateness of the different favorably received responses in the way it chooses between gestures. 

Specifically, we designed a simple generic method for converting the user preferences gathered during the action-response elicitation study into a probabilistic behavior model that determines which action the robot should perform based on which gesture it just detected from the user. The equation we crafted to transform the average ratings into probabilities is as follows:
\begin{equation}\label{eq:3}
p_{g|a} = \frac{(\max(r_{g|a} - \eta, 0))^m}{\sum_{i = 1}^{N}{(r_{i|a}-\eta})^m} 
\end{equation}
where $p_{g|a}$ is the probability with which the robot will perform the specified gesture $g$ given the user action $a$, $r_{g|a}$ is the average user rating that gesture $g$ received when presented as a robot response to user action $a$, $\eta$ is the neutral value on the rating scale, $m$ is a positive power that controls how strongly higher-rated gestures should be favored (chosen to be 3 for our study), $N$ is the total number of gesture options being considered (usually 4 for our study), and $r_{i|a}$ is the average user rating that gesture $i$ received when presented as a robot response to user action $a$. This formula subtracts the neutral value $\eta=5$ from each average rating to focus only on responses that were positively received; if a rating is below neutral, the value of zero is used instead to ensure the robot never performs that gesture in response to this user action. 

The numerator and denominator terms in equation~\eqref{eq:3} are raised to the power of $m=3$ to increase the probability the robot will select highly rated responses; other powers could be used to provide other blends between exploration of different options and exploitation of the known best choice. For example, $m=1$ sets the gesture probabilities to be directly proportional to the relative strength of their received ratings, which we found to be too random. In comparison, $m=0$ would perform the $N$ gestures with equal probability (pure exploration with no dependence on user rating), and $m=\infty$ would always select the highest-rated gesture (pure exploitation with no variety). Regardless of the value of $m$, the $N$ resulting probabilities necessarily sum to unity. 

In practice, we pre-compute the contingent gesture probabilities from the average ratings shown in Fig.~\ref{fig:BehaviorMatrix}. The robot's software then implements this probabilistic behavior algorithm by generating a random number between 0 and 1 each time a response is required. The probabilities relevant to the detected action are stacked, and HuggieBot 3.0 executes the gesture corresponding to the generated number.

\subsection{Responding to Discrete Active Gestures}
\label{subsec:discrete}
The intra-hug gestures of rubbing and patting both consist of discrete hand motions that the user can perform a single time or repeat many times in a row. Thus, the robot's response to these gestures is designed to also follow a discrete action paradigm. When the perceptual pipeline detects the start of a new user rub or pat, the robot first determines which discrete gesture to respond with by applying equation~\eqref{eq:3} to the average user ratings from the appropriate row (rub or pat) of Fig.~\ref{fig:BehaviorMatrix}. Because hold received a positive average rating in response to user rub actions, the robot chooses between all four gestures in this situation with probabilities of $p_{\textrm{hold|rub}}=0.01$, $p_{\textrm{rub|rub}}=0.30$, $p_{\textrm{pat|rub}}=0.14$, and $p_{\textrm{squeeze|rub}}=0.55$. In contrast, hold received a slightly negative average rating in response to user pat actions, so the hold response is never chosen after a pat is detected ($p_{\textrm{hold|pat}}=0.00$); the robot's three non-zero gesture response probabilities in this case are $p_{\textrm{rub|pat}}=0.27$, $p_{\textrm{pat|pat}}=0.21$, and $p_{\textrm{squeeze|pat}}=0.52$. The gestures with which the robot responds are all executed with fixed timing, as in the action-response elicitation study. 

The overlapping time windows used in our perception algorithm cause many successive detection events to be triggered over the course of a single long user action. Thus, while the robot is executing the response to a detected discrete active gesture, it enters a state of ignoring new detected actions for 2.5 seconds (slightly longer than the time it takes the robot to perform the selected gesture). This way, the robot does not accumulate a backlog of queued actions it needs to respond to, which could result in a never-ending pat or rub response. It should be noted that even in this state, the robot constantly checks to see if the user has indicated a desire to end the hug by either releasing pressure on the back chamber or leaning back against the arms, as shown in Fig.~\ref{fig:robot_state_diagram}. If either of these conditions occurs, the state-machine overrides whatever gesture the robot is performing and begins releasing the user. Otherwise, the robot returns to responding to new detected actions after the delay elapses. 

\subsection{Responding to Modal Active Gestures}
\label{subsec:modal}
When a new squeeze is detected, HuggieBot 3.0 calculates the response gesture probabilities by substituting the average user ratings from the squeeze row of Fig.~\ref{fig:BehaviorMatrix} into calculating equation~\eqref{eq:3}.  Because hold received a slightly negative average rating in response to user squeezes, the hold response is never chosen after a squeeze is detected ($p_{\textrm{hold|squeeze}}=0.00$); the robot's three non-zero gesture response probabilities in this case are $p_{\textrm{rub|squeeze}}=0.10$, $p_{\textrm{pat|squeeze}}=0.09$, and $p_{\textrm{squeeze|squeeze}}=0.81$. If the chosen gesture is a discrete rub or pat, the response proceeds as described in the previous Section.  However, the squeeze response to a squeeze action merits special handling because of its fundamentally different nature and the popularity of this robot gesture among users.

Unlike the repeated discrete motions of rubs and pats, squeezes are modal active gestures that move between two states that apply lower and higher pressure on the partner, respectively. As shown by the action-response elicitation study, user squeeze actions can vary greatly in duration, and the start and end of a robot squeeze are quite salient to the user. To avoid repeatedly squeezing the user for fixed time intervals, and to follow the user suggestion to have the robot's extended squeeze duration match theirs, HuggieBot 3.0 responds somewhat differently to squeezes compared to discrete active gestures. Specifically, if the robot's response to a detected squeeze is to squeeze the user back, as soon as the robot detects a squeeze, it enters a squeezing state, and it leaves this state only when the perception algorithm detects a hold (which means the chamber pressure has decreased to near the baseline value and that the microphone is not detecting strong activity). At this point, the robot stops squeezing and returns to the normal hug state, where it reacts to detected gestures in the normal way.

As shown in Fig.~\ref{fig:robot_state_diagram}, the robot's response to user squeezes has one other special aspect. Observing user behavior during the action-response elicitation study showed that several users squeezed the robot at the same time that they performed rubs or pats. Thus, when the robot is in the squeezing state, it continues monitoring the detected gestures. It stays in the squeezing state when the perceptual algorithm continues detecting user squeezes.  If a rub or a pat is instead detected, it responds to this discrete active gesture layered on top of a modal active gesture in a way similar to usual discrete active gestures. The robot's response is again determined using the probabilistic behavior paradigm given in equation~\eqref{eq:3}. Because the robot is already in the squeezing state, the robot squeeze response is not considered, so the robot chooses between the remaining response options for the detected action, which are all discrete and thus take a fixed amount of time. In this way, a user can squeeze the robot, be squeezed in response, then rub or pat the robot's back, and receive a rub or a pat in response, still during the squeeze.

\subsection{Responding to Passive Gestures}
\label{subsec:passive}
The final case happens when the perceptual pipeline detects that the user is passively holding the robot and not performing an active gesture. Based on the positive user ratings and comments found for proactive robot gestures in the action-response study, it is important for hugging robots to occasionally take the initiative to perform a gesture when it detects only hold gestures from the user. Providing proactive affective touch is more subtle than responding to discrete actions because the robot must take the temporal element into account to ensure that the person is genuinely not doing anything over an extended period. To determine the frequency at which proactive robot affective touch should occur, we looked at the rate at which the investigator remotely activated the gestures in the action-response elicitation study. The average time delay between the end of one proactive robot gesture and the beginning of the next one across the holding hugs of the 32 participants was 1.5 seconds. Thus, the behavior algorithm waits until it detects the hold user action detection for 1.5 seconds in a row (approximately seven overlapping windows) before proactively initiating an intra-hug gesture by the robot. Our probabilistic behavior model is again used to determine which gesture the robot should perform, as described in equation~\eqref{eq:3}; the calculated conditional probabilities are $p_{\textrm{hold|hold}}=0.11$, $p_{\textrm{rub|hold}}=0.22$, $p_{\textrm{pat|hold}}=0.10$, and $p_{\textrm{squeeze|hold}}=0.57$. This definition completes our description of how HuggieBot 3.0 behaviorally responds to gestures detected by its perceptual pipeline.

%% file: sections/07_validation.tex
\section{Validation Study -- Methods}
\label{Validation}
This study aims to test and validate the new platform of HuggieBot 3.0 (Section~\ref{Improvements}), its perceptual pipeline for detecting and classifying intra-hug gestures (Section~\ref{Detection}), and its probabilistic behavior algorithm for responding to detected gestures (Section~\ref{Response}). Regarding the platform improvements, we specifically sought to evaluate the user response to the updated hug initiation process (Section~\ref{subsec:initiation}) and the updated robot hand placement based on estimated user height (Section~\ref{subsec:initiation}). The validation study we conducted was similar to the action-response elicitation study except for the key difference that the robot was always behaving autonomously. The Max Planck Society's Ethics Council approved all methods for this study under a new framework agreement protocol number, F011B. The investigator recruited voluntary participants in the same manner as described in Section \ref{UserStudyMethods}. 

\subsection{Participants} 
The sixteen participants were English-speaking volunteers recruited from Stuttgart, Germany. No participants were employed by the Max Planck Society, so all were compensated at a rate of 8 euros per hour. While the COVID-19 pandemic reduced the number of participants we were able to safely recruit, we believe the rated quality of the resulting hugs and the qualitative feedback provided by these diverse users were more important to the study than the total quantity of hugs exchanged. This study was every user's first time interacting with any version of HuggieBot and the first robotic interaction of any kind for most users. Half of our participants were men, and half were women; 93.75\% were non-technical. The average height reported by our users was 1.69~m, with a standard deviation of 0.10~m. The participants ranged in age from 22 to 38 (mean = 30, standard deviation = 4.76) and came from ten different countries. Over 80\% of the participants identified as enjoying receiving hugs from others. 

\begin{figure}[t]
\includegraphics[width=\columnwidth, trim = {2cm 12cm 2cm 12cm},clip]{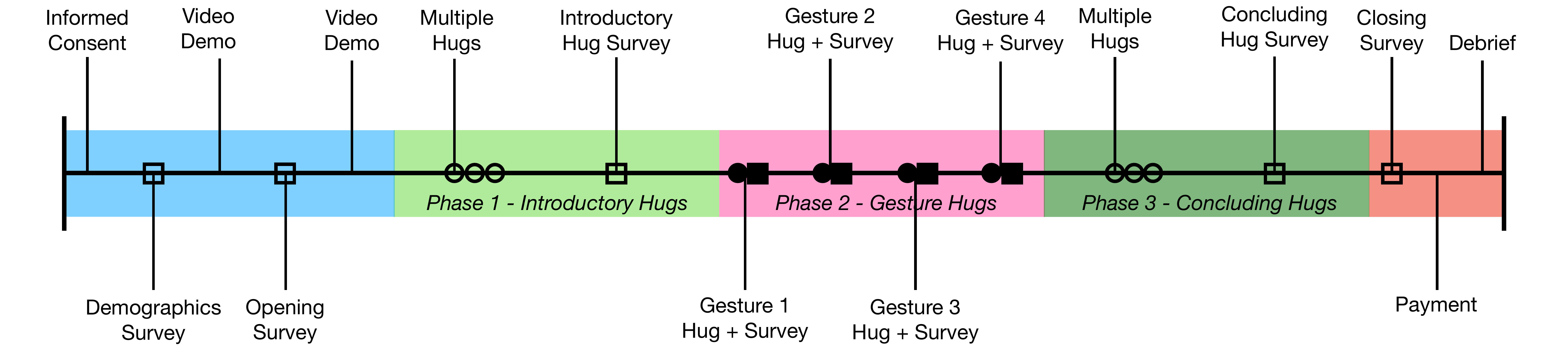}
\vspace{-0.3cm}
\caption{The timeline of the validation study. The colored boxes represent the five phases of the experiment. The left-most vertical line marks the start of the user study, and the right-most vertical line marks the end. Vertical lines coming off the center timeline show the order of the user's activities. Squares on the timeline indicate when the user filled out a survey, while circles indicate when they performed hugs. If the shape (square or circle) is not filled, that activity was performed in an open-ended way. If it is filled in with black, the activity was more controlled. Note that each user chose the number of open-ended hugs to perform in phases 1 and 3 rather than always performing the three hugs indicated in the timeline.}
\label{fig:ValidationTimeline}
\vspace{-0.3cm}
\end{figure}

\subsection{Procedure} 
Like the action-response elicitation study, prospective users were required to confirm their eligibility to participate by email, given the exclusion criteria and the local COVID-19 regulations. The timeline for this study can be found in Fig.~\ref{fig:ValidationTimeline}. Once the recruit arrived at the experiment site, the investigator explained the study. They read over and signed the informed consent document and video release form after asking the experimenter any questions. At this point, the experimenters started recording the study and preparing the robot while the user filled out the demographics survey. 

Instead of verbally explaining how to interact with HuggieBot 3.0, the investigator showed the user a video on a laptop of a sample user hugging the robot; the video demo is included as supplementary material for this article. Without providing any verbal instructions, this video shows the user how the robot asks for a hug (lifting its arms and saying ``Can I have a hug, please?''). It also shows how to prompt the robot to close its arms (walking forward) and how to cause the robot to release (either by releasing pressure off the robot's back or by leaning back against the robot's arms). We chose this new standardized method of introducing users to the robot to more closely mimic how a user might learn how to hug the robot if they encountered it in the wild. We intentionally did not show the user performing any gestures on the robot in this instructional video to avoid biasing users toward this method of human-robot interaction. After watching this video, the user filled out an opening survey to document their initial impressions of the robot (Table \ref{table:OpenCloseSurvey}). Importantly, this study included no formal practice hugs. Before the user began to interact with the robot physically, they watched the 30-second-long instructional video one more time as a refresher, since some pilot participants had forgotten what to do while completing the opening survey. The experimenter prompted them to ``pay particular attention to the timing between the user and the robot.''

\begin{table}
\caption{The fifteen questions asked in the opening and closing surveys of the validation study. Users rated their answers on a continuous sliding scale from ``disagree'' (0) to ``agree'' (100) with a resolution of 0.1.}
\vspace{-0.2cm}
\label{table:OpenCloseSurvey}       
\begin{small}
\begin{tabularx}{\linewidth}{lX}
\hline\noalign{\smallskip}
I feel understood by the robot\\
I trust the robot\\
Robots would be nice to hug  \\
I like the presence of the robot\\
I think using the robot is a good idea \\
I am afraid to break something while using the robot\\
People would be impressed if I had such a robot\\
I could cooperate with the robot \\
I think the robot is easy to use\\
I could do activities with this robot\\
I feel threatened by the robot \\
This robot would be useful for me \\
This robot could help me\\
This robot could support me\\
I consider this robot to be a social agent\\
\noalign{\smallskip}\hline
\end{tabularx}
\end{small}
\vspace{-0.3cm}
\end{table}

This study included three phases of physical interaction with HuggieBot 3.0; the robot always behaved autonomously, with no intervention by any experimenter and no changes to the program it was executing. The first phase was a natural hugging scenario. The users were given no additional instructions on how to hug the robot. They could perform as many hugs as they wanted, they could position their arms any way they wanted, and they could choose whether or not to perform any intra-hug gestures. They were simply told to hug the robot naturally several times, with a minimum of two hugs. We started the study with this open-ended phase to observe how users would naturally interact with the robot and to see whether they would discover the robot's ability to detect and respond to intra-hug gestures without experimenter prompting. Once they were satisfied with their introductory hugs, users were asked to fill out a single short survey for all the hugs they had performed, including answering the questions found in Table~\ref{table:FourQuestions}. These four questions were asked about \textit{the entire hug experience}, which included the hug initiation, the embrace, the hand location, any intra-hug gestures they may have encountered or triggered, and the release. 

\begin{table}
\caption{The four questions asked after the introductory and concluding hug sessions (phases 1 and 3) in the validation study.}
\vspace{-0.4cm}
\label{table:FourQuestions}       
\begin{small}
\begin{tabularx}{\linewidth}{lX}
\hline\noalign{\smallskip}
This robot behavior seemed (Unnatural -- Natural)\\
These hug interactions felt (Awkward -- Enjoyable)\\
These hugs made the robot seem (Socially Stupid -- Socially Intelligent)\\
These hugs made the robot seem (Unfriendly -- Friendly)\\

\noalign{\smallskip}\hline
\end{tabularx}
\end{small}
\vspace{-0.3cm}
\end{table}

The second phase of the experiment included four hugs with intra-hug gestures conducted in a somewhat controlled manner.  Before each hug, the user was instructed to perform a single gesture during the hug after the robot's arms had closed. They could perform the specified gesture as many times as they liked. The four gestures they were asked to perform were hold, rub, pat, and squeeze. We used an $4 \times 4$ balanced Latin square to counter-balance any effects of presentation order \cite{LatinSquare} and recruited participants in multiples of four to have an equal number of participants in all the presentation orders. After each hug, users were asked to fill out a short survey about the robot behavior they had just experienced in response to their gestures. This survey started with a free-response question asking for the user's ``first impressions of this interaction.'' Then, they were asked to mark all the robot responses they experienced (options were robot arms staying still, robot arm moving vertically, robot arm tapping on back, and robot arms tightening hold). Finally, the participant used a sliding scale from 0 (hate) to 10 (love) to rate how they felt about the robot's response to the action they performed during the hug. Before moving on to the next part of the study, the user verbally explained their rating to the experimenter. 

In the third phase of this study, users were once again asked to hug the robot naturally. They were allowed to hug the robot as many times as they liked. Afterward, users answered the same short questionnaire from the first phase, including the questions found in Table~\ref{table:FourQuestions}.

At the end of the third phase, users filled out a closing survey. The closing survey included the same questions from the opening survey (Table \ref{table:OpenCloseSurvey}) plus questions asking users to rate the quality of the four different robot gestures they experienced during the experiment (hold, pat, rub, squeeze), as also done at the end of the action-response elicitation study. There were also two questions aimed at seeing how users responded to two new features we implemented: users rated the naturalness of the hug initiation method and the appropriateness of the robot's hand placement. Finally, users could provide additional free-form comments at the end.

%% file: sections/08_results2.tex
\section{Validation Study -- Results}
\label{Results2}
We analyzed the system's ability to perceive intra-hug gestures, the users' survey responses about the quality of their interactions, and user comments from all parts of the validation study to characterize HuggieBot 3.0's skill at autonomously and interactively hugging users. For reference, two annotated videos of participants performing active intra-hug gestures during their concluding hugs are included as supplementary material for this article; the annotations indicate the actions performed by the user, the actions perceived by the robot, and the gestures that the robot decides to execute in response. A video of a participant performing an extended hold during a hug in phase 2 is also included, annotated in the same way. Appendix \ref{app:vidplots} presents plots showing the joint angles, joint torques, microphone signal, and pressure signal for all three of the annotated videos included as supplementary material.

\subsection{Performance of the Perception Pipeline}
\begin{figure}[t]
\includegraphics[width=0.8\columnwidth, trim = {0.1cm 0.1cm 0.1cm 0.1cm},clip]{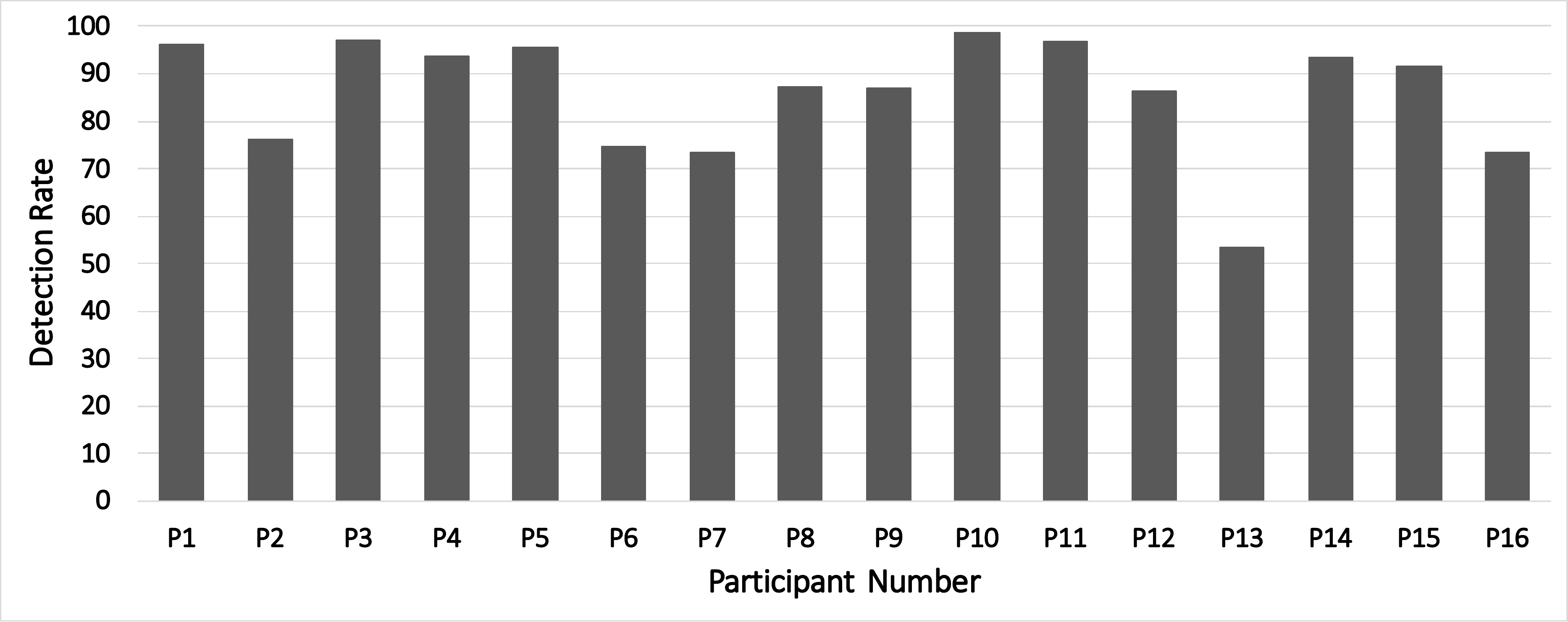}
\vspace{-0.3cm}
\caption{Gesture perception accuracy per participant.}
\label{fig:validation_accuracy}
\vspace{-0.3cm}
\end{figure}

We analyzed data recorded by the robot during the second phase of the validation study to estimate the gesture perception pipeline's accuracy. The recorded data for each hug included timestamps, microphone voltage, and pressure values with a 45~Hz sampling rate, as well as the gestures detected every 10 samples. To determine accuracy, one of the authors visualized the data from the four hugs in the second phase of the experiment for all users (a total of 64 trials). As previously described, the participant was instructed to perform a specific gesture (e.g., squeeze) during each of these hugs. Since the users often interleaved their active intra-hug gestures with passive pauses, we calculated accuracy as the number of data points detected with the correct gesture or hold over the total number of detected gestures between the start and end of each hug. The mean detection accuracy for the sixteen participants was 85.9\% (standard deviation = 12.5\%). This overall accuracy is comparable to the perception pipeline's accuracy (88\%) on the data set collected in the action-response elicitation study. Figure~\ref{fig:validation_accuracy} presents the perceptual  accuracy for each participant. The gesture detection accuracy is above 86\% for eleven participants, above 73\% for four other participants (P2, P6, P7, P16), and at 53.5\% for one participant (P13).

\begin{figure}[t]
\includegraphics[width=\columnwidth, trim = {0cm 2.75cm 0cm 2.75cm},clip]{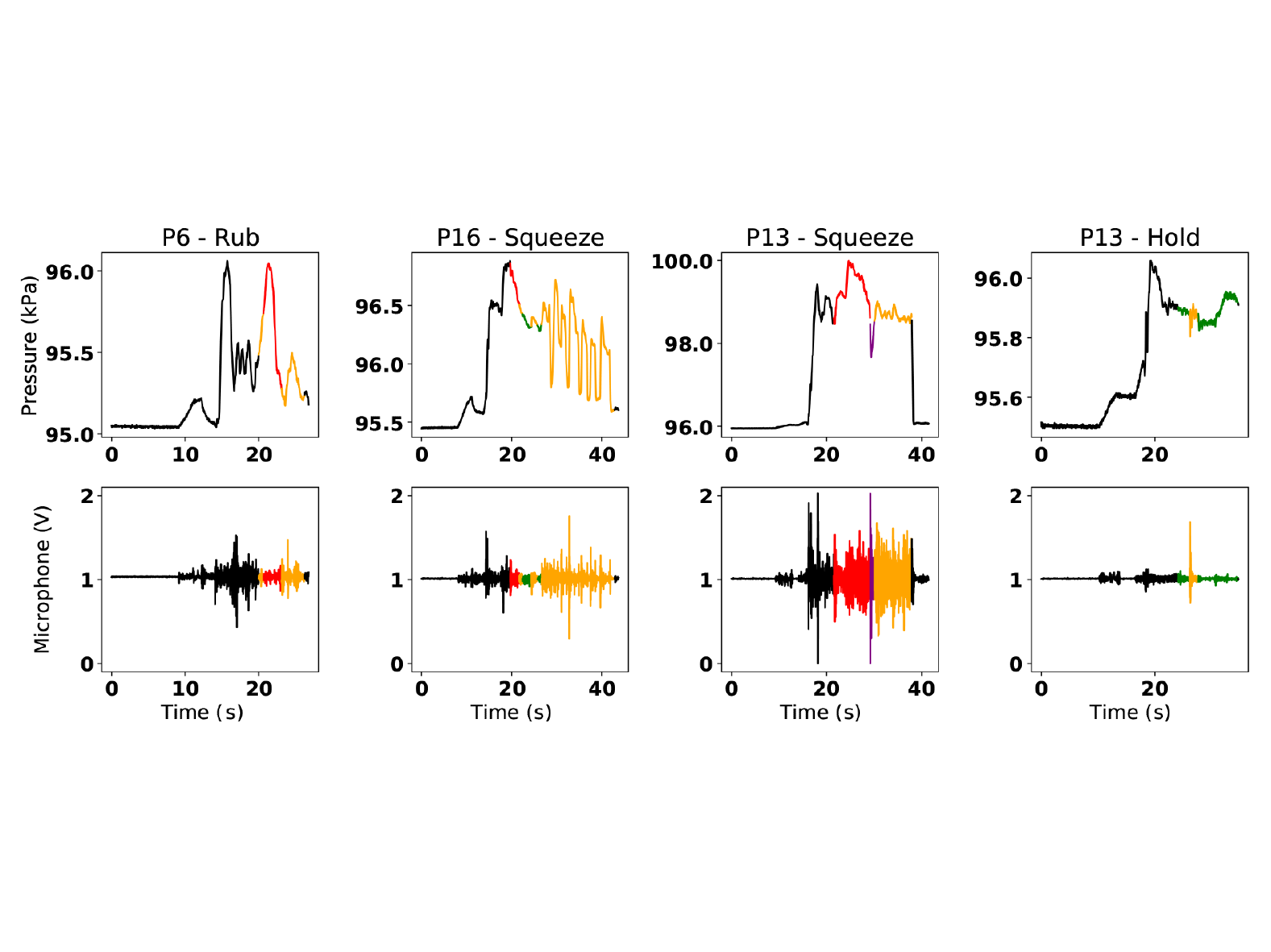}
\vspace{-0.3cm}
\caption{Example misclassified trials where the participants performed a gesture in an unexpected or uncommon way. Each panel's title gives the participant number and the action they were instructed to perform. The colored data points mark the time periods automatically classified as each of the four intra-hug gestures; following the color convention of Fig.~\ref{fig:teaser}, hold is green, rub is yellow, pat is purple, and squeeze is red.}
\label{fig:WrongTrials}
\vspace{-0.3cm}
\end{figure}

We examined the trials that had low detection rates and noted that sometimes the participants performed a gesture in an unexpected way. For example, for the squeeze gesture, P16 released the pressure on the chamber and applied it again instead of increasing the pressure (P16 -- Squeeze, Figure \ref{fig:WrongTrials}). At other times, the participant applied little pressure (P13 -- Squeeze, Figure \ref{fig:WrongTrials}) or squeezed the robot while performing another gesture (P6 -- Rub, Figure \ref{fig:WrongTrials}). In some cases, the participant's accidental move or gesture, such as shifting their body against the front of the robot's torso, was detected as a rub (P13 -- Hold, Figure \ref{fig:WrongTrials}). The algorithm also sometimes misclassified the start or end of a pat as rubbing. 

\subsection{User Experience}
For all statistical analyses of the user responses to the questionnaires, we used an alpha value of $\alpha = 0.05$ to determine significance. To handle the problem of multiple comparisons and to lower the likelihood of a type I error, we used the Holm-Bonferroni method for alpha correction \cite{Holm1978}. The Bonferroni correction is the simplest and most conservative approach to apply to multiple comparisons; however the consequence is an increased likelihood of type II errors \cite{vanderweele2019some, weisstein2004bonferroni}. When the Bonferroni correction is considered too conservative, it is recommended to use a Holm-Bonferroni or a Hochberg correction \cite{armstrong2014use}. We have chosen a Holm-Bonferroni correction because it has the added benefit of showing more power compared to the Bonferroni procedure \cite{kim2015statistical}.  
When comparisons are significant, we report effect sizes using MATLAB 2019b's built-in function for Pearson's linear correlation coefficient: \texttt{$\rho$ = corr(X)}, where X is the matrix of data being evaluated. The value of $\rho$ signifies the strength of the bivariate relationship. $\rho = 0.1$ shows a small effect size, $\rho = 0.3$ indicates a medium effect size, and $\rho = 0.5$ or above signals a large effect size. 

\subsubsection{Opening and Closing Surveys}
\begin{figure}[t]
\includegraphics[width=\columnwidth, trim = {7cm 0cm 5cm 0cm},clip]{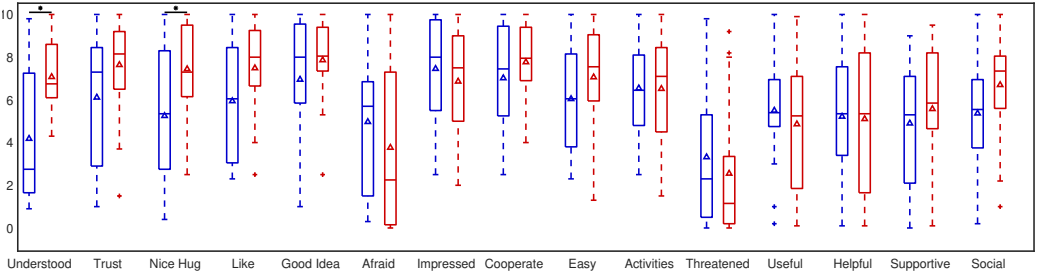}
\vspace{-0.3cm}
\caption{Box plots comparing the responses to the opening (blue) and closing (red) survey questions given in Table~\ref{table:OpenCloseSurvey}. The top and bottom of each box represent the 25th and 75th percentile responses, respectively, while the line in the center marks the median, and the triangle shows the mean. The lines extending past the boxes show the farthest data points not considered outliers. The + marks indicate outliers. The two black lines with stars at the top of the graph indicate statistically significant differences.}
\label{fig:PrePost}
\vspace{-0.3cm}
\end{figure}

Box plots of the user responses to the opening and closing survey questions from Table \ref{table:OpenCloseSurvey} are shown in Fig. \ref{fig:PrePost}. In this study, answers were submitted on a continuous sliding scale from 0 (disagree) to 10 (agree), so a paired t-test comparison of the opening and closing survey was conducted after verifying the data were normally distributed. After applying the Holm-Bonferroni method for alpha correction (with a $n = 15$ for the fifteen questions in the survey), we found that users felt significantly more understood by the robot ($p = 0.0023$, $\rho = 0.21$) and felt the robot was significantly nicer to hug ($p = 0.0035$, $\rho = 0.59$) after the experiment. Three other comparisons approached significance: users liking the presence of the robot ($p = 0.0084$, $\rho = 0.72$), thinking the robot could support them ($p = 0.0372$, $\rho = 0.92$), and viewing the robot as a social agent ($p = 0.0208$, $\rho = 0.68$). 

\subsubsection{Introductory and Concluding Hugs (Phases 1 and 3)}
\begin{figure}[t]
\includegraphics[width=\columnwidth, trim = {4cm 0cm 4cm 0cm},clip]{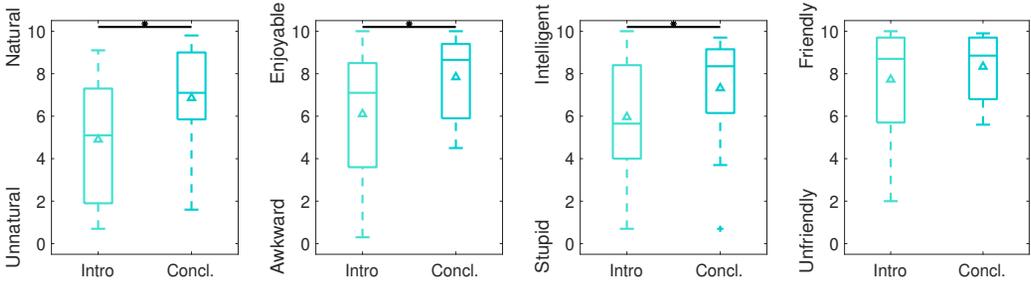}
\vspace{-0.3cm}
\caption{A box plot comparison of the responses to the four survey questions about the introductory hugs (light turquoise) and concluding hugs (dark turquoise), as listed in Table~\ref{table:FourQuestions}. The three black lines with stars indicate statistically significant differences.}
\label{fig:IntroConcl}
\vspace{-0.3cm}
\end{figure}

The first and third phases of the study involved asking the participants to perform several natural hugs with the robot. The responses to the four questions from Table~\ref{table:FourQuestions} asked after the introductory hugs (phase 1) and the concluding hugs (phase 3) can be seen in Fig.~\ref{fig:IntroConcl}. These responses were also analyzed using a paired t-test comparison and a Holm-Bonferroni alpha adjustment. We found that users did not initially find the robot's hugging behavior very natural, but their opinion of it significantly improved by the end of the study ($p = 0.0138$, $\rho = 0.47$). At the end, users also found the robot hugs significantly more enjoyable ($p = 0.0028$, $\rho = 0.77$). Finally, after hugging the robot repeatedly, users found the robot significantly more socially intelligent ($p = 0.0049$, $\rho = 0.84$) than their initial impressions. The users did not significantly change their opinion about the robot's friendliness over the course of the study; it was already rather highly rated after the introductory hugs.  

\begin{figure}[t]
\includegraphics[width=\columnwidth, trim = {3cm 0cm 3cm 0cm},clip]{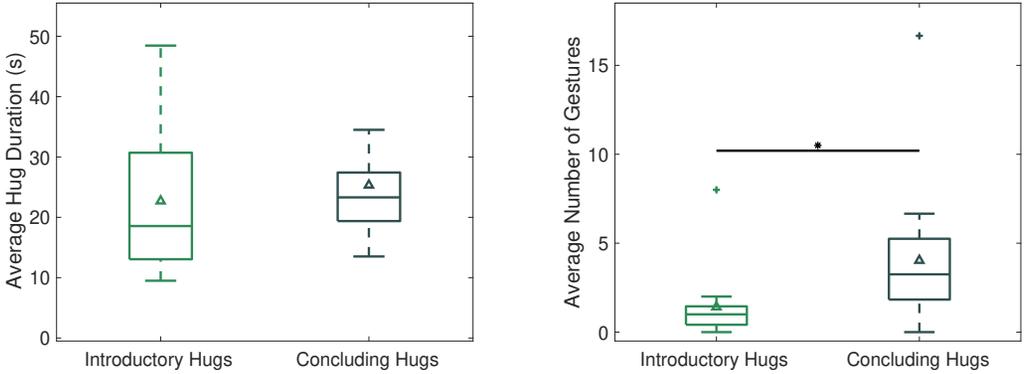}
\vspace{-0.3cm}
\caption{A box plot comparison of the average duration (left subplot) and the average number of intra-hug gestures (right subplot) of the hugs during the introductory hug phase (pale green) and the concluding hug phase (dark green) of the validation study. The black line with a star indicates a statistically significant pairwise difference.}
\label{fig:AverageDuration}
\vspace{-0.3cm}
\end{figure}

We also investigated the average hug duration and the average number of gestures users performed during the introductory and concluding hug phases (Fig.~\ref{fig:AverageDuration}). For the introductory hug phase, the average hug duration was 22.7$\pm$12.6 seconds, where 12.6 seconds is the standard deviation. The concluding hug phase had an average hug duration of 25.3$\pm$11.0 seconds. The average number of active user gestures (rub, pat, squeeze) detected during the introductory hug phase was 1.41$\pm$1.84. The average number of active gestures detected during the concluding hug phase was 4.03$\pm$3.95. We ran a paired t-test on the average hug duration and the average number of gestures in these two phases. Although twelve of the sixteen users engaged in longer hugs with the robot during the concluding phase than during the introductory phase, we did not find a significant difference for hug duration. However, we did find a significant difference for the number of gestures performed ($p < 0.001$, $\rho = 0.89$), with significantly more gestures performed during the concluding hugs. 

\subsubsection{Gesture Hugs (Phase 2)}
\begin{figure}[p]
\includegraphics[width=\columnwidth, trim = {0cm 0cm 0cm 0cm},clip]{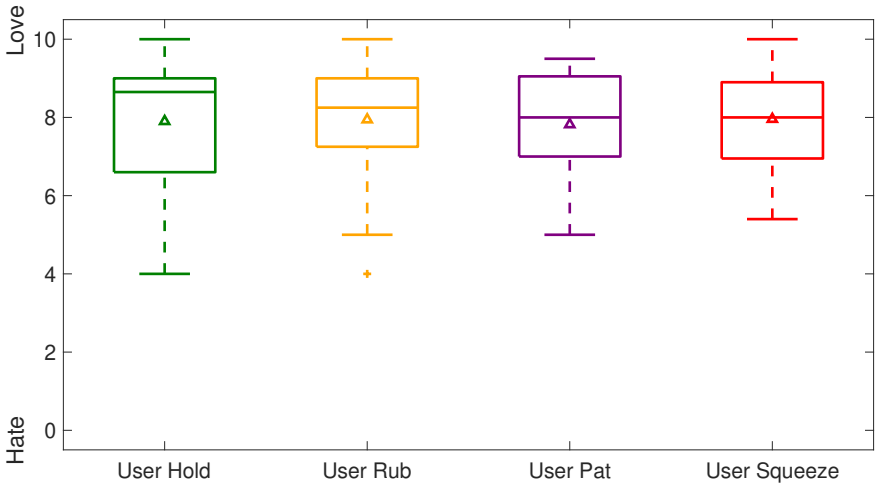}
\vspace{-0.3cm}
\caption{A comparison of the user ratings of the robot's autonomous responses to the four different intra-hug actions performed by users in the second phase of the validation study.}
\label{fig:RobotResponseRating}
\vspace{-0.3cm}
\end{figure}

Users experienced four hugs during phase two of the study,  performing a specific gesture as many times as they liked in each hug. The robot used its perceptual pipeline and probabilistic behavior algorithm to autonomously decide how to respond to each of the user's actions. Figure \ref{fig:RobotResponseRating} shows a box plot of the user ratings for the overall robot responses for each performed user action. For all user actions, the average rating of the robot's combined response is around eight out of ten with a standard deviation of around 1.5 ($r_{\textrm{hold}}=7.90\pm1.77$, $r_{\textrm{rub}}=7.94\pm1.64$, $r_{\textrm{pat}}=7.82\pm1.35$, and $r_{\textrm{squeeze}}=7.96\pm1.42$), indicating that the robot's responses were perceived very positively by users.

\begin{figure}[p]
\includegraphics[width=0.8\columnwidth, trim = {5cm 15cm 0cm 15cm},clip]{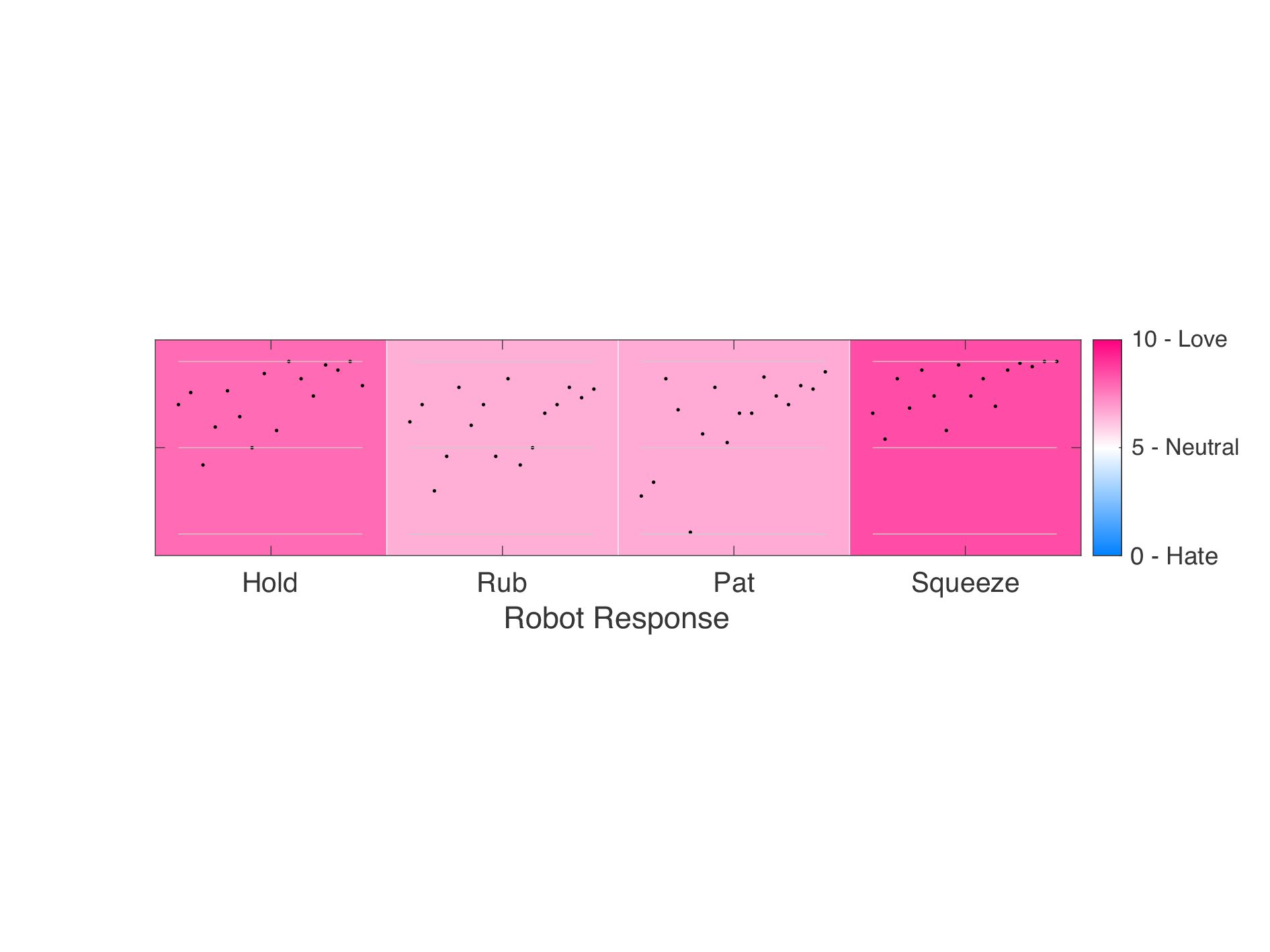}
\vspace{-0.3cm}
\caption{A matrix showing the user ratings of the quality of the various robot responses from the validation study, following the visualization approach used in Fig.~\ref{fig:BehaviorMatrix}.}
\label{fig:QualityMatrix2}
\vspace{-0.1cm}
\end{figure}

\begin{figure}[p]
\includegraphics[width=\columnwidth, trim = {4cm 0cm 4cm 0.5cm},clip]{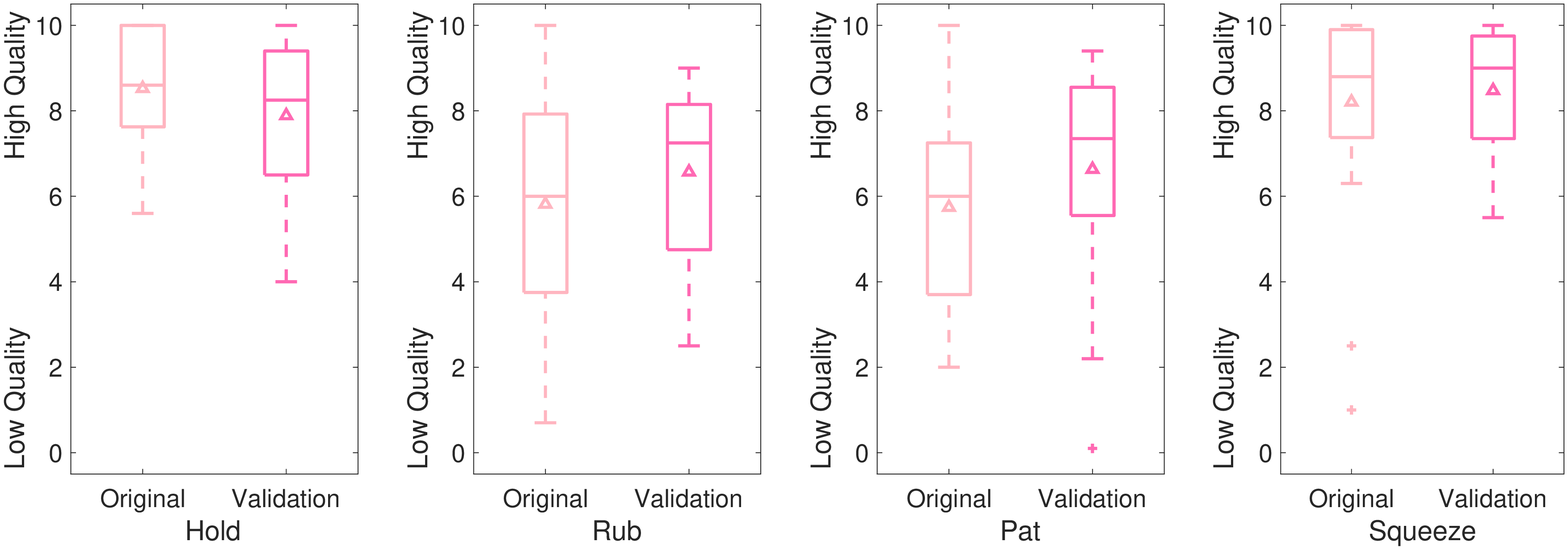}
\vspace{-0.3cm}
\caption{A comparison of the responses to the quality of the robot's gestures during the original action-response elicitation study (pale pink, Fig.~\ref{fig:QualityMatrix}) and the validation study (dark pink, Fig.~\ref{fig:QualityMatrix2}).}
\label{fig:QualityComparison}
\vspace{-0.3cm}
\end{figure}

\subsubsection{Additional Closing Survey Questions}
At the end of the experiment, users were asked to evaluate the quality with which the robot performed the four different intra-hug gestures. This question was presented and framed the same way as in the action-response study, for continuity and comparison. A matrix showing the average rating on a color scale and each user's individual rating in dots can be seen in Fig.~\ref{fig:QualityMatrix2}. Figure~\ref{fig:QualityComparison} shows side-by-side comparisons of the gesture quality ratings obtained in the original action-response elicitation study and those from the validation study.  We then compared the quality ratings of the four different robot actions between the two studies using unpaired t-tests. We had a different number of participants and an entirely separate population. A Holm-Bonferroni alpha adjustment with $n=4$ did not show any significant differences in the perceived quality of the robot gestures.

\begin{figure}[t]
\includegraphics[width=\columnwidth, trim = {2.5cm 0cm 3cm 0cm},clip]{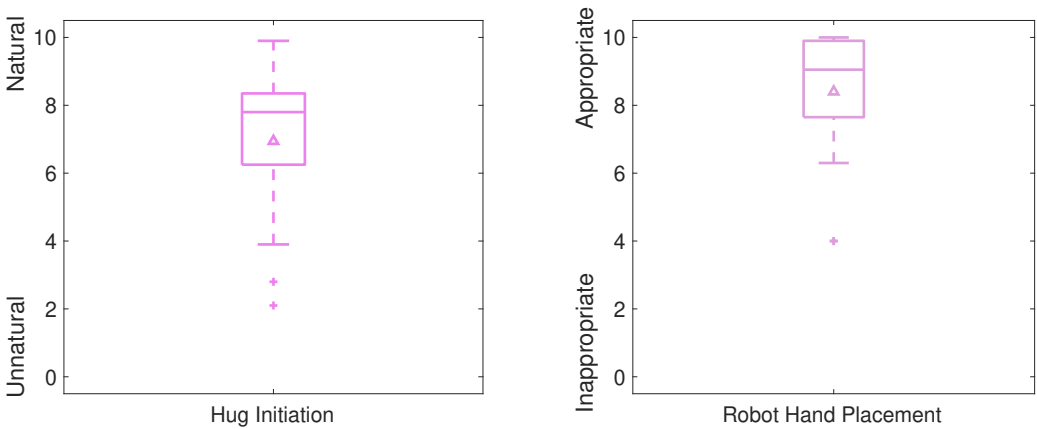}
\vspace{-0.3cm}
\caption{Box plots showing user ratings of the naturalness of the hug initiation process (left) and the appropriateness of the robot hand placement (right).}
\label{fig:TriggerPlacement}
\vspace{-0.3cm}
\end{figure}

Fig. \ref{fig:TriggerPlacement} reports how the users rated the naturalness of the hug initiation process and the appropriateness of the robot's hand placement on their back.  The average rating of the hug initiation was 6.94$\pm$2.22 on a scale from unnatural (0) to natural (10). The average rating of the robot's hand placement was 8.40$\pm$2.02 on a scale from inappropriate (0) to appropriate (10). One outlier rated robot hand placement a 4 out of 10. This user was one of the tallest users we tested, with a height of 1.83~m. Two other users had the same height as this user and rated it 6.3 and 9 out of 10. This user tended not to stand straight when the robot was estimating his height, and thus, HuggieBot 3.0 thought he was shorter than he was, which resulted in the robot's arms occasionally being placed lower than the user would have preferred.

\subsection{User Comments} 
Once again, some of the most informative results come straight from the users in the form of free-response written and verbal comments. Half of the users commented explicitly on the naturalness and ease of initiating the hug with the robot. 31.25\% of the sixteen users specifically commented that the ``timing was well synchronized/done perfectly on time'' (P10, P13) between initiating a hug and the arms closing. One user (6.25\%) initially felt the arms closed too slowly. However, by the end of the experiment, this user stated the timing was ``natural and comfortable'' (P12) for them. Interestingly, two users (12.5\%) thought the robot's arms closed slightly faster than with people and felt they had to walk slightly faster than usual. Regarding the robot's hand placement, 75\% of users mentioned that the location of the robot's hand was ``good'' and ``well-placed exactly where [they] want it'' (P15) at all times. Users specifically mentioned that at no point did it make them feel uncomfortable, which is an improvement from the user comments from the action-response elicitation study. 

After the introductory hugs, six users (37.5\% of sixteen) mentioned that hugging the robot felt ``strange'' (P2) or ``weird'' (P6, P7, P9) because it was their first time interacting with a robot, and they were not sure what to expect or how the robot would respond. While these users initially found hugging a robot strange, that does not mean they did not enjoy their experience. 50\% of all the users mentioned enjoying their first group of robot hugs, particularly commenting on the robot's ``warmth'' (P4, P5, P10, P12, P16), ``friendliness'' (P2, P7, P12), and ``comfortable appearance'' (P2, P4, P6, P7). Some users even went as far as to say ``I enjoyed the experience as I am away from my family and a warm hug is always comforting'' (P4). Only three users (18.75\%) mentioned feeling uncomfortable or nervous initially. 

After the concluding hugs session, user comments were much more positive. Fifteen out of our sixteen participants (93.75\%) shared positive comments mentioning ``great overall experience'' (P3, P4) and that they ``liked the robot's responses'' (P4, P5, P6, P12, P15, P16). Users mentioned that they found the hugs to be both ``comfortable'' (P2, P4, P6, P7) and ``natural'' (P4, P7 P10). The one user who did not have as positive comments as the rest mentioned that ``the experience did not feel natural, but it was fun to test how a robot can interact with a human in this way'' (P8). Three users who were initially tentative about the robot in their comments about the introductory hugs stated how comfortable they eventually found the robot (18.75\%). Six users (37.5\%) were ``amazed'' (P3, P7) by the experience and came to think of the robot as a ``friend'' (P2, P7, P14) by the end of the concluding hugs phase. 

We also asked users to share their opinions about the overall experience at the end of the entire study. Many users (62.5\%, ten out of sixteen) shared that they felt the entire experiment was a ``great experience'' or that they ``loved hugging the robot'' (P5, P6). Two users (12.5\%) mentioned that while they enjoyed the interaction, they did not understand the purpose of such a robot as they felt that human hugs are ``irreplaceable'' (P12, P13). Several of our users (37.5\%, six of sixteen) mentioned how valuable they found this robot especially given the current COVID-19 pandemic, mentioning that ``emotional and mental health are also important'' (P7) but are often forgotten or not addressed. 

%% file: sections/09_discussion.tex
\section{Discussion and Limitations}
\label{Discussion}
\subsection{Discussion}
This research project investigates the previously unstudied phenomenon of intra-hug gestures during hugs between a human user and an autonomous robot. We found users to be positively interested in hugging a robot that can both respond to their gestures and proactively perform gestures on its own. The results support all six new of the design guidelines we proposed for hugging robots.  

\paragraph*{G4: Hug Initiation} First, our revised version of the fourth tenet states that when a hugging robot is the one initiating the interaction, it should autonomously initiate a hug when it detects a user in its personal space by inviting the user for a hug; it should then wait for the user to begin walking toward it before closing its arms to ensure a consensual and synchronous hugging experience, as done by HuggieBot 3.0. The previous evaluation of HuggieBot 2.0 did not find a statistically significant user preference between a hug initiated with a button press or a hug initiated via computer vision detection of the user's approach~\cite{TheSixHugCommandments}. Users gave our new two-phase hug initiation method an average naturalness rating of almost a 7 out of 10. All users were able to initiate hugs with the robot after watching a simple video demonstration twice, without requiring the detailed instructions that were previously provided. Thus, we conclude that the new hug initiation method is an improvement over the previously tested methods. We also had pilot-tested an alternative three-step hug initiation method that was harder for users to master and received an average naturalness rating of only 4.62 out of 10 from twelve pilot participants. Our observations repeatedly indicate that consensual and synchronous hug initiation is indeed important. Although it still has room for improvement, rather than forcing the user to hurry up and wait, the new two-phase method more closely mimics how hugs occur between humans and lets the user decide when to start hugging the robot. Relevantly, \citet{walters2008human} found that both the voice used and previous experience interacting with the robot can have an effect on the mean approach distance to a robot. \citet{mead2017autonomous} extensively worked on understanding interaction potential based on human-robot proxemics. These prior works also support the need for consensual and synchronous hug initiation highlighted by G4.

\paragraph*{G7: Adaptation to User Height} The next new design guideline states that a good hugging robot should also perceive the user's height and adapt its arm positions to contact the user at appropriate locations. We thus extended our platform's perceptual capabilities beyond detecting a user's approach so that HuggieBot 3.0 attempts to embrace the user at the proper height -- not too high, and not too low.  Our newly developed height adjustment system was created in direct response to user feedback about the inappropriateness of HuggieBot 2.0's hand placement in the action-response elicitation study. The average rating for the appropriateness of HuggieBot 3.0's hand placement's was an 8.4 out of 10, with several users giving it the highest rating possible; these ratings indicate that the proposed approach usually succeeds at adjusting for user height. We also believe that the non-significant increases in the quality ratings for the rub and pat gestures (which we did not adjust other than the location of the robot's left hand) can probably be attributed to this improved placement. In human-human touch interactions, there are relationship-specific maps of body regions where touch is considered appropriate \cite{suvilehto2015topography}; the areas where contact is allowed increase with the emotional bond to the toucher. Therefore, from human-human touch interaction research, we confirm that our robot's hand placement on the user's back is important to avoid taboo zones and ensure the comfort of all users, regardless of the relationship they associate with the robot. Thus, we conclude that users prefer hugging a robot that adjusts its arm placement to match their height.

\paragraph*{G8: Gesture Perception} The next design guideline centers on enabling a hugging robot to accurately and reliably detect and classify user gestures applied to its torso in real time, regardless of the user's hand placement. Both of our user studies demonstrated the excellent haptic sensing capabilities of the pressure sensor and microphone inside the inflated chamber of HuggieChest; simple signal-processing and machine-learning techniques were able to detect and classify contacts very well, even when the pipeline was transferred to new sensing hardware and adapted with limited new training data. As our subjects used various hand positions on the robot (both arms below the robot's arms, both above, or one above and one below), we found that this non-localized haptic sensing system works well regardless of user hand placement on the back chamber. Based on the average gesture detection accuracy of 86\% for the sixteen participants in the validation study, along with the positive opinions users shared when the robot responded to their gestures in both studies, we believe the results support the validity of this design guideline. In further exploring cases when the detection algorithm did not perform well, we found that users frequently performed the gesture in an unexpected or uncommon way; in some cases, these variations may have come from the user's limited vocabulary for intra-hug gestures in English. As a surprising benefit, however, the perception pipeline was able to detect some rubs and pats performed at the same time as a squeeze, even though it had not been trained to do so.

\paragraph*{G9: Fast Response} The ninth design guideline simply states that users like a robot that responds quickly to their intra-hug gestures. As seen in Fig.~\ref{fig:BehaviorMatrix}, when a user performed an intra-hug gesture on the robot's back, and the robot did not respond, users perceived this as a neutral robot behavior on average. In their written and verbal comments to the experimenter, users indicated they did not feel like the robot ``understood'' them, ``knew [they were] there,'' or ``wanted to support/comfort [them].'' Users clearly preferred when the robot indicated that it knew the user had performed an intra-hug gesture and responded quickly in some way. When we were piloting the validation study, we noticed that a technical error occasionally caused the sampling rate of the microphone and pressure sensor to drop to about half the normal value.  We found that when users interacted with the robot in this condition, the delay was highly noticeable and detracted from the user experience. Several pilot subjects mentioned that ``it felt like the robot was performing random actions at random intervals, not in response to anything I was doing.'' During our final validation study, when the sampling rate issue was fixed, the robot responded quickly to user actions, and our subjects were delighted. They repeatedly performed the same gesture to experience the robot response again. Users would often comment to the experimenter during the hug itself, saying ``it tapped me back!'' (P4), or remarking after the hug that ``every time I did an action, it noticed and did something back to me!'' (P16) because they were so pleasantly surprised at the responsiveness of the robot. This user desire for a fast robot response time aligns with expert therapist opinions that robotic systems for perceiving social touch from humans also have strict timing requirements \cite{BurnsSeifiLeeKuchenbecker2021}. A relevant common psychology study is called ``infant response to still-face,'' where a mother, who had been interacting and playing normally with her child, suddenly stops smiling and talking to her infant while the experimenters observe the child's response \cite{toda1993infant}. Children commonly become distressed when their mothers no longer respond, and can cry, become fussy, and grasp at themselves and their mothers, trying to get attention. The \textit{delayed response} from their mother significantly upsets them. Through the combination of results from our two user studies, as well as research from human-human interactions, we believe there is support for our ninth guideline suggesting a fast robot response to intra-hug gestures.

\paragraph*{G10: Response Variety} The next design guideline states that hugging robots should adopt a gesture response paradigm that blends user preferences with slight variety and spontaneity. When starting this project, we believed that hugging robots should always reciprocate the same intra-hug gesture the user had performed. The results from the action-response elicitation study surprised us by showing that rote reciprocation is not expected and would not be perceived in a fully positive way. If users preferred gesture reciprocation, we would see a dark pink diagonal in \ref{fig:BehaviorMatrix}. Instead, we see a slight preference for a robot to respond to any user action with a squeeze. Speaking with our users showed us that they appreciate variety in robot responses. Something about the unpredictability of the response leads users to feel it is more ``alive.'' Users also mentioned that having the robot respond with the same action as the user performed feels ``too mechanical,'' because based on the input you know exactly what output you will receive. The results from the action-response elicitation study thus support this design guideline, as do the very positive user reactions to the resulting robot behavior algorithm tested in the validation study. We believe the slightly spontaneous robot hugging behavior enabled by our simple probabilistic behavior paradigm (equation~\eqref{eq:3}) succeeds at blending user preferences with spontaneity to reasonably match natural human exchanges of intra-hug gestures. The behavior algorithm's tendency to prefer exploration versus exploitation can also easily be adjusted by changing the value of the exponent $m$. Interestingly, most human-human research found that mimicry increases perception of another person \cite{chartrand2013antecedents, van2009love}. The chameleon effect is a common phenomenon in which humans unconsciously mimic the gestures and facial expressions of an interaction partner to match their type of social expression and level of extroversion \cite{chartrand1999chameleon}. In this instance, our users made it clear that explicit mimicry from the robot is not appreciated, but that they did want similar levels of support. This finding could be seen as similar to human interactions. In a social environment, it would be uncomfortable if a social partner was obviously copying you, but responding to gestures with similar levels of enthusiasm seems warranted. 

\paragraph*{G11: Proactive Robot Gestures} This design guideline states that hugging robots should occasionally provide unprompted proactive affective social touch to the user through intra-hug gestures. The findings of~\citet{ClinicalRobotTouch} made us initially hypothesize that users would dislike robot-initiated affective social touch delivered via unprompted intra-hug gestures; their users reacted negatively when a robot attempted to comfort them by touch but did not mind functional contact from the robot. The findings of our two studies explicitly contradict this hypothesis and support G11. We were so surprised by these ratings during the action-response elicitation study that after the user had finished explaining their ratings, the experimenter asked the follow-up question, ``so just to clarify, it did not bother you that you did nothing and the robot unprompted started rubbing/patting/squeezing you?'' Users confirmed that not only did they not mind this robot behavior, but they also \textit{enjoyed and appreciated it}. Users indicated that while in the other cases, the robot would respond to their gestures, here, they felt the robot was comforting them. In these cases, many users commented that they felt the robot's emotions and feelings and that it cared more about them when it chose on its own to perform a gesture, rather than just responding. Although more work needs to be done to confirm this positive finding, it seems that appropriately framed robot-initiated affective touch may be key to creating robots that can provide good emotional support to human users.

How can we grapple with the seemingly conflicting findings between our work and \citet{ClinicalRobotTouch}? We believe these results are not as different as they may appear. The users in our studies \textit{agreed to enter into a hug with a robot}, so we believe they also felt at least partially responsible for initiating the affective touch that occurred during the resulting hug. Once this initial boundary is broken, we believe users are more receptive to proactive robot affective touch, for example, a rub, pat, or squeeze. Users in all of our studies have appreciated that HuggieBot \textit{politely asked} them for a hug, thereby allowing them to agree to this affective touch. Many users even responded affirmatively to the robot every time it asked the question, even though they knew it never listened to their answer. By changing the hug initiation method to be prompted by the robot lifting its arms for a hug and asking ``Can I have a hug, please?'' and then waiting with its arms outstretched for the user to approach, we further put the initiation of the affective social touch on the user, solidifying that it is their choice to enter the hug. We believe user initiation is key to acceptance of future social, affective touch from a robot. We therefore firmly believe G11's statement that robots can evoke user feelings that the robot is alive and caring by occasionally providing unprompted affective touch to the user, as delivered by HuggieBot 3.0 through intra-hug gestures.

\subsection{Limitations}
While the research described in this paper presents several key contributions to robotic hugging and broader social-physical human-robot interaction, we nevertheless acknowledge several limitations of our work. 

The first limitation is the somewhat artificial methodology of our studies. We recognize the importance of conducting in-the-wild studies for human-robot interaction research. By conducting these studies in a laboratory environment, we have a self-selection bias of our participant pool. Only users who were interested in hugging a robot chose to participate in the reported studies. Unfortunately, due to the current COVID-19 pandemic, lab studies were the safest way to conduct research on hugging robots. We were able to screen participants for potential health risks and thoroughly sanitize the robot between subjects. For the validation study, we changed the robot's introduction from verbal instructions from the experimenter to having the user watch a simple video of another user hugging the robot. This video introduction was meant to mimic how users would learn to use the robot in the wild. Once the COVID-19 crisis has ended in our region, we look forward to conducting a thorough in-the-wild study to see how many everyday people would and would not be interested in hugging a robot. Additionally, the COVID-19 pandemic also reduced the number of participants we could recruit for our second user study; the results from the validation study have lower statistical power (roughly 50\%) than would typically be presented and thus should not be the only results taken into consideration.

Next, our refined fourth guideline is limited in that it addresses only situations where the robot is the one initiating the interaction. If the user is the one requesting the hug from the robot, the first half of the guideline should be ignored, and the robot should observe only the second half, which is ``waiting for the user to begin walking toward it before closing its arms to ensure a consensual and synchronous hugging experience.'' 

Additionally, other researchers such as \citet{ClinicalRobotTouch} have shown the importance of context with respect to the acceptance of social touch. The reported studies did not use any specific context beyond the narrative descriptions provided by the experimenter. Future work could study HuggieBot in different contexts, such as a nursing home or shopping mall, and evaluate the effect each context has on user expectations and interpretations of the interaction. 

We have also identified two main limitations of the Kinova JACO arms used in both HuggieBot 2.0 and 3.0. While we adjusted the hug initiation process to accommodate this limitation better, the Kinova JACO arms just cannot move fast enough to mimic the speed of a human's arms closing. These arms were selected for safety reasons, and this speed limit was considered during that choice. After extensive user testing, we have found that the speed limit causes a real limitation on the naturalness of the user experience because subjects have to wait several seconds before the arms have fully closed around them. A second limitation of these arms is that repeated small movements of the first, second and third joints (shoulder lift, shoulder pan, and elbow flex, respectively) occasionally result in a sudden short but fast jerking movement, which startles the user. This phenomenon occurs only rarely during repeated rubs or pats. When this issue occurred during the two reported studies, we immediately commanded the robot to release the user, checked that they were okay, verified whether they wanted to continue, discounted the trial with the malfunction, and restarted the trial. Because the sudden motion is very small, this technical glitch never hurt a user. However, it is likely that it negatively affected some user ratings of HuggieBot as a whole, as well as the robot's ability to perform rubs and pats. 

This work is also limited by the fact that we have simplified the problem of gesture classification significantly by focusing on only four gestures. There are infinitely many gestures a person could choose to perform during a hug, and there are infinitely many ways they could perform each gesture. A user could even combine multiple gestures together. We chose to select a simple subset of four classic gestures and their combinations (e.g., squeeze-pat) as a first step into this new research area. We currently do not estimate the intensity with which users perform these gestures, nor do we measure the location where gestures are performed. Interesting future steps would be to measure the intensity and/or the location of user gestures to enable the robot to reciprocate gestures with an appropriate intensity and/or location on the user’s back. 

Another limitation is that both of our studies asked the user to perform intra-hug gestures somewhat artificially. After placing their hands on the robot's back, users had to wait for the robot's arms to close fully before performing a gesture. This pause was used to collect baseline measurements for the microphone and pressure signals so that the real-time perception pipeline could determine what gestures the user subsequently performed. We found that many users naturally wanted to start performing the gesture immediately after beginning the hug, regardless of the robot's arm movements. To collect data and then test our algorithm's accuracy, we also asked users to perform only one gesture per hug, though they could perform the gesture repeatedly if they chose. This restriction was also somewhat unnatural. We found that many users naturally wanted to combine gestures. We added the natural hug scenarios in phase 1 and phase 3 of the validation study to address this limitation. 

Our action-response elicitation study challenged users with the difficult task of separating the appropriateness of the robot's response from the quality with which HuggieBot 2.0 performed the gesture. We had them explain their ratings to the experimenter to ensure they understood the distinction and were answering the question correctly. Nonetheless, the robot's gesture quality probably affected other user ratings. Users experienced a similar challenge in the validation study, where we again asked them to separately rate the robot responses from the quality of the robot's gestures; gesture quality also thus probably affected these results. Another limitation from the action-response elicitation study is that we did not ask users to rate the naturalness of the hug initiation process or the appropriateness of the robot's hand placement. We had not realized these aspects of HuggieBot 2.0's behavior would garner negative comments and thus need to be adjusted for the new version of the platform. Thus, we had to rely on written and verbal comments to evaluate the effects of these changes.

An additional limitation involves our evaluation of the user experience. Though we aimed to assess it in an accurate manner and specifically collected data in multiple ways to facilitate comparisons, it is possible there were still problems. First, whenever collecting self-reported data, the questions will be subject to the interpretation of the users, who may not have the same understanding \cite{podsakoff1986self}. We used pilot testing to make our questions as clear and unambiguous as possible. We also did our best to conceal which aspects of the robot we were evaluating, so as to avoid the demand effect, where a participant tries to respond in a way to confirm or deny the hypothesis of a study \cite{nichols2008good}. Though we acknowledge that a participant could have deduced what we were testing for, we do not think it is likely that participants responded untruthfully because we consciously conveyed equipoise throughout the experiment and because several of our findings did not match our initial hypotheses. Finally, as with any technology, there is the concern of the novelty effect, that users' attitudes and preferences will wane over time \cite{leite2009time, kidd2005human}. We aimed to mitigate this effect by conducting long experiments and querying user opinions both before and after one and a half hours of robot hugs. Nevertheless, for almost all of our users, the reported study was their first experience interacting with HuggieBot, and for many of them it was their first time interacting with any robot. To better evaluate the influence of the novelty effect on user evaluations with HuggieBot, future studies should have users interact with HuggieBot over the course of many weeks or months.

Finally, having a robot that fully understands a human hug is very challenging. We acknowledge that the current version of our robot does not deliver on the full aspirational goal of a hugging robot. Rather, HuggieBot simulates a hug in a reasonably compelling way, and our data suggest that users enjoy the hug and can engage with the robot and relate to it as an autonomous being. However, in its current state, HuggieBot does not have an internal emotional model similar to humans, and thus it is not capable of engaging in the embodied emotional experience of a hug. 


%% file: sections/10_conclusion.tex
\section{Conclusion and Future Work}
\label{Conclusion}
This article started by collecting a large data set that shows the characteristic microphone and pressure signals for 32 diverse users performing four intra-hug gestures (hold, rub, pat, and squeeze) on the inflated torso of HuggieBot 2.0. We used these recordings to create a perception pipeline that detects and identifies these different gestures in real time. Ratings and comments in reaction to how the robot responded showed that users do not want a robot that mimics their gestures back to them; instead, they want a robot that responds quickly and naturally to their gestures with some level of unpredictability similar to the choices made by a human hugging partner. We thus developed a behavior algorithm that uses conditional probabilities based on user ratings to determine how our robot should respond after detecting a particular user action during a hug, distinguishing between discrete active gestures, modal active gestures, and passive gestures. We also made several critical changes to our robot platform, including changing the method of hug initiation, adjusting the robot's arm positions to the estimated height of the user, constructing a new robot torso, and improving the quality of the robot's embrace. We tested this upgraded version (HuggieBot 3.0) together with its new perceiving and acting skills on a new set of sixteen diverse users who had not previously interacted with this robot. Users were generally very pleased with the robot's responses to their actions. The platform changes seemed to improve the quality of the interaction, and the perception and behavior approaches that we developed worked very well. Therefore, we conclude that hugging robots should be able to perceive and respond quickly to intra-hug gestures from users.

While there are several possibilities for future work related to this project, we are most interested in five future directions. First, we want to deploying a future version of HuggieBot in a pedestrian environment to measure how many people would naturally be interested in interacting with such a robot in the wild. We will first need to improve the robot's approaching user detection algorithm, which currently works well with only one user in the frame. To function in a non-controlled (non-lab) environment with many passersby potentially in the robot's field of view, we will need to make this aspect of the robot more robust. HuggieBot will also need to detect if a user changes his/her mind and decides not to hug the robot, rather than eternally waiting with its arms raised, as HuggieBot 3.0 does. Another challenge of conducting an in-the-wild study will be dealing with non-haptic noise detected by the microphone. We are currently taking a baseline at the start of every hug to determine the mean microphone output, but we are not estimating the magnitude or spectrum of background noise that needs to be filtered out. We found that nearby construction sometimes caused high levels of ambient noise during pilot testing, causing our perception pipeline to mistakenly think that the user was continuously rubbing the robot's back. 

A simpler improvement to the robot's behavior centers on reducing the likelihood that users accidentally end the hug before they intend to do so. We noticed that several users in both studies briefly took their hands off the robot's torso to adjust their grip and grasp the robot tighter when performing a squeeze, as shown in one of the annotated videos included in the supplementary materials. Because the chamber pressure decreased, the robot assumed the user wanted to be released, so it ended the hug. To make our system more robust, rather than act on a single low pressure value, if the pressure decreases and increases again quickly, the robot should not release the user because they probably do not want the hug to end. Given the negative social impact of not letting go of a user who wants to be released, it is important to find the right balance between fast and reliable hug termination and avoidance of accidental user releases.

A third aspect we are interested in investigating is to see if the calming effects of robot hugs are somewhat physiologically similar to the calming effects of human hugs. We propose to investigate this question by safely inducing stress upon voluntary participants and providing either an active human hug, a passive human hug, an active robot hug, or a passive robot hug. Over the course of the experiment, we would periodically collect saliva samples from users to measure the cortisol and oxytocin levels in their bodies, also recording heart rate, video, and subjective opinions of the experience. Running a study of this type would enable us to more confidently say whether an embodied affective robot like HuggieBot 3.0 truly has the potential to supplement human hugs in situations when requesting this form of comfort from others is difficult or impossible. 

Another element we are looking forward to researching is the extent to which a future version of HuggieBot can help strengthen personal relationships between people who are separated by a physical distance. To do this, we have already developed a mobile app, the HuggieApp, that allows remote users to send customized hugs to local users (via the robot). The sender can customize the hug's duration and tightness, and they can add a variety of intra-hug gestures. They can even replace the robot's animated face with a video message for the receiver. The local user redeems the hug by scanning a custom QR code on their mobile phone at the robot's camera, which is located above its face screen. Users can re-redeem their favorite hugs as many times as they want, which could be especially meaningful if the original sender has passed away. For several months, we plan to observe pairs of platonic users interacting with each other through this future version of HuggieBot; we hope to evaluate the relationship's perceived closeness to see the extent to which this type of embodied affective robots can help bridge physical distance between people.

A final potential future direction would to be transfer the capabilities of HuggieBot to a more generalized companion or care robot. Of particular interest could be combining the comfort of a hug with socially assistive robots in care homes. Adults over 65 are the fastest growing demographic, and there are not enough workers to care for them \cite{roberts20119}.
Depression in older adults is extremely common as they can struggle with medical illness, cognitive dysfunction, physical separation from their friends and family members, or a combination of the three \cite{taylor2014depression}. Robots like Paro \cite{Chang2013} and Stevie \cite{taylor2021exploring} show that not only is there great interest in robots being used with older adults, but that they can be beneficial to these users. Considering the proxemic work of \citet{mead2017autonomous} could help us enable HuggieBot to perform both robot- and human-initiated hugs, making the entire experience more natural and hopefully more enjoyable for users. Equipping a more generalized care robot, which can already assist with a variety of tasks, with the ability to provide high-quality embraces could help address older adults' unmet social, physical, and emotional needs.

%% file: sections/11_appendix.tex
\appendix


\section{Data Plots For Supplemental Videos}
\label{app:vidplots}
The supplemental material for this article includes three annotated videos showing how the detection and classification algorithm and the behavioral response algorithm work in real time; the videos show footage recorded from three different participants during the validation study presented in the article. This appendix provides annotated plots of the robot's joint angles, joint torques, microphone signal, and pressure signal for the corresponding videos. 

\begin{figure}[tp]
\includegraphics[width=\columnwidth, trim = {1cm 5cm 1cm 5cm},clip]{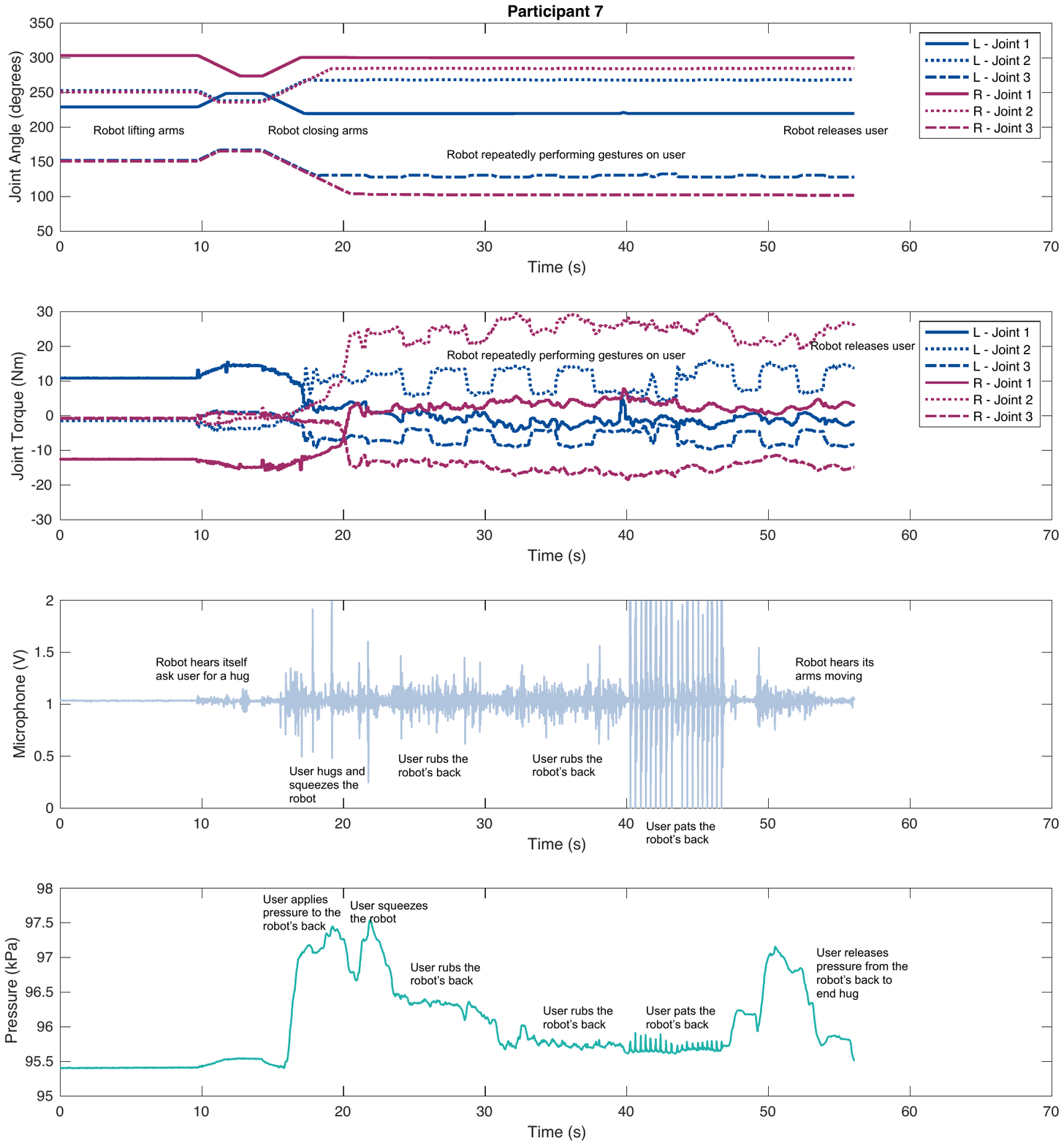}
\vspace{-0.3cm}
\caption{Participant 7.}
\label{fig:P7_annotated}
\vspace{-0.3cm}
\end{figure}

\begin{figure}[tp]
\includegraphics[width=\columnwidth, trim = {1cm 5cm 1cm 5cm},clip]{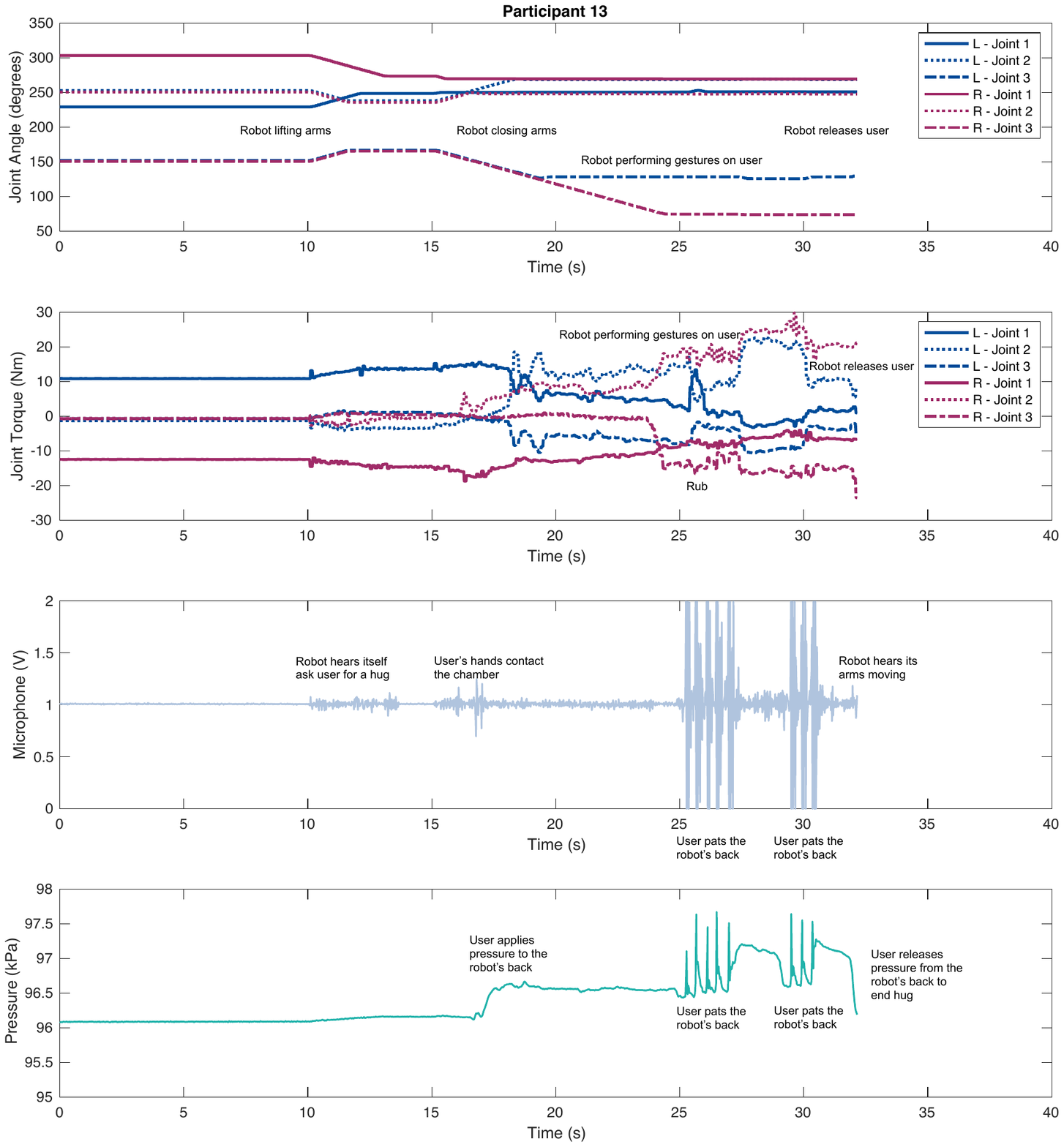}
\vspace{-0.3cm}
\caption{Participant 13.}
\label{fig:P13_annotated}
\vspace{-0.3cm}
\end{figure}

\begin{figure}[tp]
\includegraphics[width=\columnwidth, trim = {1cm 5cm 1cm 5cm},clip]{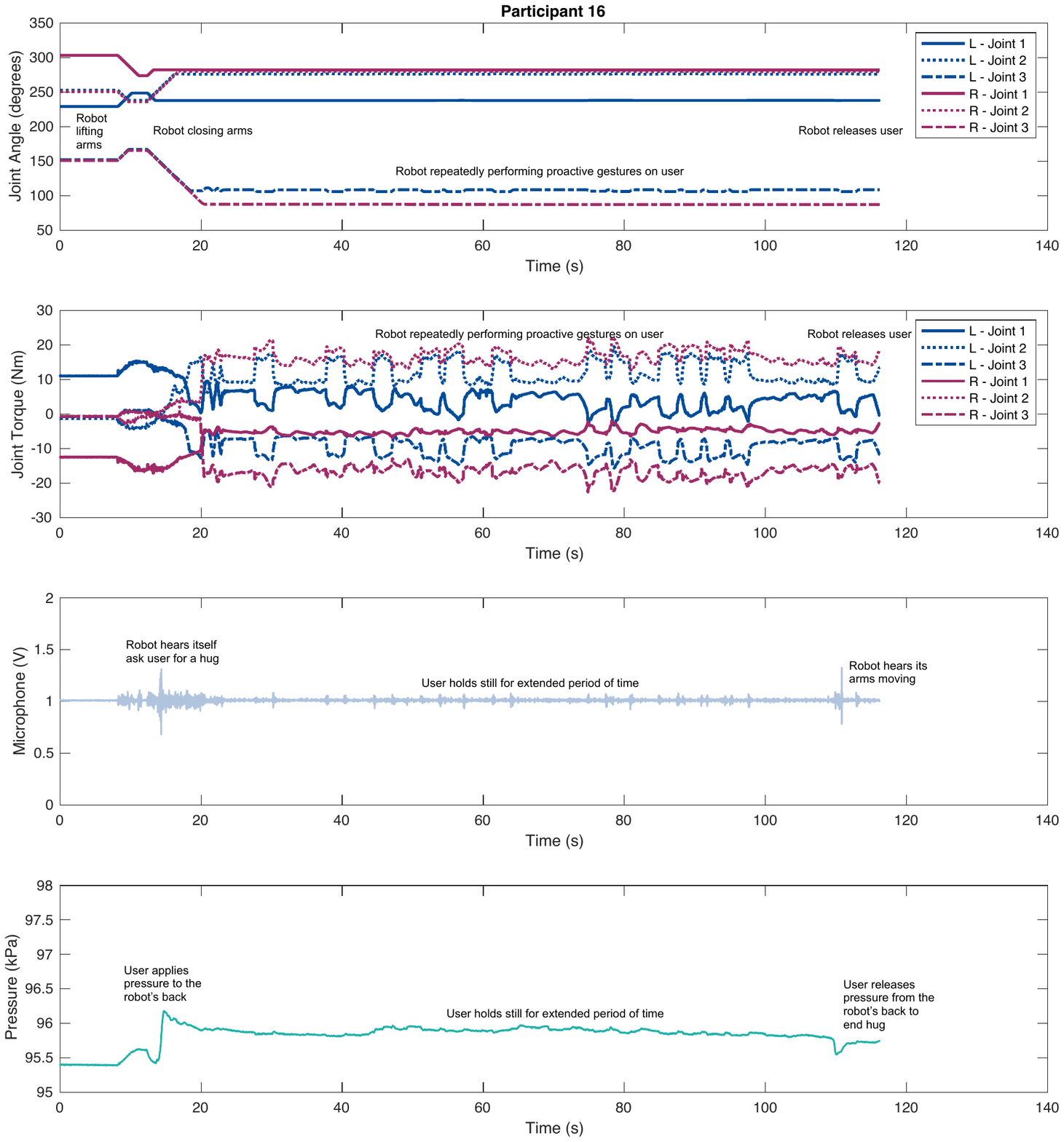}
\vspace{-0.3cm}
\caption{Participant 16.}
\label{fig:P16_annotated}
\vspace{-0.3cm}
\end{figure}